\pdfoutput=1

\documentclass[11pt]{article}
\usepackage{booktabs}

\usepackage{ACL2023}

\usepackage{times}
\usepackage{latexsym}
\usepackage[most]{tcolorbox}
\usepackage{listings}
\usepackage[T1]{fontenc}

\usepackage[utf8]{inputenc}

\usepackage{microtype}
\usepackage{makecell}

\usepackage{inconsolata}

\usepackage{xcolor}
\usepackage{times}
\usepackage{latexsym}
\usepackage{latexsym}
\usepackage[normalem]{ulem}
\usepackage{microtype}
\usepackage{inconsolata}
\usepackage{booktabs}
\usepackage{multirow}
\usepackage{graphicx}
\usepackage{amsmath,amssymb}
\usepackage{enumitem}
\usepackage{subcaption}
\usepackage{siunitx}
\usepackage{xspace}
\usepackage[most]{tcolorbox}
\tcbuselibrary{raster}
\usepackage[utf8]{inputenc}
\usepackage{textgreek}

\usepackage{tabularx}
\usepackage{ragged2e}

\usepackage{amssymb}
\tcbset{
  mypromptbox/.style={
    colback=gray!3,
    colframe=black!20,
    fonttitle=\bfseries,
    breakable,
    enhanced,
    boxrule=0.5pt,
    arc=1mm,
    left=1mm, right=1mm, top=1mm, bottom=1mm
  }
}
\tcbset{
  diagbox/.style={
    enhanced,
    colback=white,
    colframe=black,
    arc=5mm,
    boxrule=1.0pt,
    left=9pt,right=9pt,top=9pt,bottom=9pt,
    halign=flush left,
    valign=center
  },
  outerdash/.style={
    enhanced,
    colback=black!6,
    colframe=black,
    arc=5mm,
    boxrule=0pt, 
    left=12pt,right=12pt,top=12pt,bottom=12pt,
    borderline={1.0pt}{0pt}{dash pattern=on 3pt off 2pt}
  },
  innerbox/.style={
    enhanced,
    colback=black!4,
    colframe=black,
    arc=5mm,
    boxrule=1.0pt,
    left=10pt,right=10pt,top=10pt,bottom=10pt,
    halign=flush left,
    valign=center
  }
}

\title{Choosing the Right Language Mode at Inference Time for Multilingual Reliability}

\author{Ekata Mitra \and Ameeta Agrawal\\
  PortNLP, Portland State University, USA \\
  \texttt{\{ekata, ameeta\}@pdx.edu} \\}

\begin{document}
\maketitle
\begin{abstract}


{Multilingual large language models often struggle to reason in low- to mid-resource languages. {Prior work has shown that translation can improve multilingual reasoning by helping models access stronger English-centric representations.} This raises a central question:  }\textit{How much translation is needed for multilingual large language models to reason reliably—and when does more translation instead trigger interference and overconfidence?} 
Using LLaMA and Qwen models, we run extensive experiments varying  \emph{text scope} and \emph{language mode} (target-only, English-only, bilingual) to evaluate both {accuracy} and {reliability}.
Our results reveal a clear trade-off: English context {often improve understanding and recover errors caused by non-English comprehension}, yet {adding redundant bilingual context intensifies} interference.~We address this {trade-off}  with {Reliability-Aware Adaptive Inference} (\textsc{RAAI}), a training-free test-time framework that (i) performs {Expected Calibration Error (}ECE)-aware routing and prompt fusion, and (ii) uses a mid-layer Risk Index (RI) to gate sequential reasoning, allocating compute only when it is likely to help and suppressing harmful bilingual redundancy.  Across {two} model families, \textsc{RAAI} enhances accuracy by {25- 37.7\% }on low-resource languages and lowers calibration error by {~3-6\%}, with the most pronounced benefits in the lowest-resource language tiers.

\end{abstract}

\section{Introduction}
While today's large language models (LLMs) have made remarkable progress in numerous tasks, a persistent gap remains between their reasoning performance in English and in other languages \cite{bandarkar2024belebele,agrawal2024evaluating,yang2023multilingual}.~These {multilingual reasoning gaps} threaten equitable deployment: models may misinterpret problem statements, fail to carry out multi-step reasoning, or produce unreliable confidence estimates in non-English settings \cite{ahuja-etal-2022-calibration,yang2023multilingual,xue2024calib}.

A common hypothesis is that these gaps stem mainly from {understanding failures}, while other errors reflect {reasoning failures} and difficulties in producing {outputs in the input language} \cite{kang2025gaps}. Multilingual LLMs appear to internally rely on an English-centric latent space (or pivot) and only later re-express decisions in the surface language.{~Such behavior can introduce errors through flawed semantic grounding or representation drift \cite{wendler-etal-2024-llamas,kojima-etal-2024-multilingual,schut2025multilingual}.~Similarly, recent interpretability work shows that LLM hidden states diverging into language-specific subspaces (rather than a truly shared space) lead to inconsistent outputs and reduced accuracy in other languages \citep{lim2025latent}, reinforcing this latent-English hypothesis.~This motivates strategies that provide English translations or bilingual prompts to encourage English-space reasoning \cite{shi2023cot,zhu-etal-2024-question}, as well as selective translation approaches that translate only when failures are detected \cite{kang2025gaps}.~{However, explicitly steering  toward an English does not uniformly yield synergy: while it promotes alignment with the model’s dominant latent space, it can also interfere with language-specific representations, particularly in bilingual contexts \cite{qi2025reason,tam2025chain}.}

Despite recent progress on bilingual prompting and selective translation \cite{shi2023cot,zhu-etal-2024-question,huang-etal-2023-languages,kang2025gaps}, we still lack a controlled understanding of how \emph{text scopes} (input fields) and \emph{language modes} (target language vs.~English translation) shape multilingual failures and reliability. In this study, we focus on parallel multilingual Multiple-Choice Question Answering (MCQA) setting, where the same question and answer options are presented across languages, enabling direct and consistent cross-lingual comparison. We introduce  a controlled $5{\times}3$ diagnostic framework on Belebele dataset \cite{bandarkar2024belebele} that varies (i)~{text scope---question-only (Q), passage-only (P), question+passage (QP), choices-only (C), or all components (All)---and} (ii) ~{language mode---target-only (T), English-only (EN), or bilingual/{Target and English (TEN)}}.
We use consistent decoding and report accuracy together with calibrated reliability signals (ECE, confidence, and entropy) following standard calibration methodology \cite{guo2017calibration,ahuja-etal-2022-calibration,xue2024calib,yang2023multilingual}.
We apply the same design on MMLU-ProX-Lite dataset \cite{xuan-etal-2025-mmlu}, using a reduced $3{\times}3$ grid (Q/C/All $\times$ T/EN/TEN).
{Our extensive analysis reveals that text scope and language mode interact in non-monotonic ways. Broader translation can improve accuracy under English prompting, yet it can also increase bilingual interference and calibration instability.


{Building on our findings, we introduce a test-time intervention method, {Reliability-Aware Adaptive Inference (RAAI)}, with two modules. The first is an \emph{ECE-aware bilingual fallback policy} that routes between prompt variants to improve reliability. The second is \emph{risk-gated sequential reasoning} which triggers additional reasoning steps  only for inputs predicted to be high-risk. Together, these components allow selective English assistance and adaptive compute at
inference time, improving multilingual reasoning accuracy and reliability while avoiding
unnecessary overhead in low-risk settings.
}

Our main contributions are:
\begin{itemize}[leftmargin=0.9em, itemsep=0.1em, topsep=0.1em]
\item A controlled analysis of text scope and language mode for parallel multilingual multiple-choice QA, with a diagnostic correctness taxonomy (understanding, reasoning, and bilingual interference).
\item {Reliability-Aware Adaptive Inference (RAAI)}: a training-free test-time method with two independent modules—bilingual fallback and sequential bilingual reasoning—for deciding when extra computation is worthwhile.
\end{itemize}


\section{Related Work}
\label{sec:related}

\paragraph{Multilingual reasoning and latent English.}
Multilingual evaluations consistently reveal an English advantage on QA and reasoning benchmarks \cite{bandarkar2024belebele,yang2023multilingual}. A growing body of work suggests that this gap may partly reflect how multilingual LLMs represent and process non-English inputs: even when the surface language is not English, internal representations often exhibit an English-centric structure \cite{wendler-etal-2024-llamas,kojima-etal-2024-multilingual,schut2025multilingual}, and unconstrained reasoning traces frequently shift toward English \cite{qi2025reason,tam2025chain}. Complementary neuron-level analyses show that multilingual processing relies on a mixture of shared and language-specific activations whose roles vary across layers, and that targeted test-time steering of these activations can improve cross-lingual transfer without changing model weights \cite{pokharel2026cross}. Motivated by this, our translation-controlled setup examines how latent English-centric processing interacts with partial translation and bilingual prompting.

\paragraph{Bilingual prompting and selective translation.}
Prior work shows that English can serve as a useful reasoning scaffold for multilingual models. Bilingual chain-of-thought \cite{shi2023cot,zhu-etal-2024-question}, cross-lingual-thought prompting \cite{huang-etal-2023-languages}, and selective translation \cite{kang2025gaps} all report gains from introducing English reasoning, particularly when it is applied selectively rather than uniformly. Consistent with this, earlier work on multilingual long-context retrieval and reasoning found that English instructions outperformed language-specific instructions in three of four non-English languages, while also showing substantial variation across languages and resource levels \cite{agrawal2024evaluating}. We build on this line of work by disentangling \emph{which} MCQA components are translated from \emph{how} English support is provided, and by using layer-aware signals to adapt inference rather than treating translation as a single fixed choice.

\begin{table*}[!t]
\centering
\small
\begin{tabularx}{\linewidth}{p{0.30\linewidth}p{0.08\linewidth}X}
\toprule
Prior work & Language Mode & Mapped Condition\\
\midrule
Target-language multilingual QA \citep{bandarkar2024layer,yang2023multilingual} & T & Question, passage, and choices remain in the target dialect throughout, with no English support at inference (our T-mode baseline)\\
English-pivot / translate-test \citep{shi2023cot,qin-etal-2023-cross,zhu-etal-2024-question} & EN & All input fields are translated into English and scored with the same decoding scheme as T, isolating the effect of full English pivoting (our EN-mode)\\
Bilingual prompting \citep{shi2023cot,zhu-etal-2024-question,huang-etal-2023-languages} & TEN & Target and English fields are provided jointly, with translation scope varied independently across Q/P/QP/C/All (our TEN-mode)\\
\bottomrule
\end{tabularx}
\caption{Mapping prior multilingual strategies onto our controlled T/EN/TEN taxonomy.}
\label{tab:baseline-taxonomy}
\end{table*}

\paragraph{Multilingual calibration and intermediate layers.}~Multilingual models are often miscalibrated, particularly for low-resource and non-Latin-script languages \cite{ahuja-etal-2022-calibration,yang2023multilingual,xue2024calib}. Prior work also finds intermediate layers can be better calibrated than final layers for multilingual inputs and that leveraging them improves ECE/Brier scores \cite{bandarkar2024layer}. We build on this by linking translation and {language} choices to layer-wise behavior and directly exploiting intermediate representations.

\paragraph{Reliability-based inference.}~Uncertainty-aware inference (e.g., abstention/refusal under low confidence) can reduce errors and hallucinations \citep{tomani2024uncertainty,kirichenko2025abstentionbench}. For multilingual settings, \citet{feng2024multilingual} uses feedback from related languages to detect knowledge gaps across dialects, whereas \citet{pokharel2025capo}  use confidence during multilingual post-training to modulate the strength of noisy preference signals. In contrast, we use calibrated confidence signals to trigger selective translation and fusion only for high-risk inputs, enabling reliability-driven routing at inference time.

We present our work alongside matched static mode baselines (T/EN/TEN) and component controls,
rather than training-based prior systems that optimize a different setup.  The matched comparison in Table~\ref{tab:baseline-taxonomy} holds the model, examples, answer space, and scoring procedure fixed while varying only the language content.

\section{Experimental Setup}
\label{sec:ex}
We describe our experimental setup in this section including the models, datasets, decoding strategy, and evaluation metrics.

\paragraph{Models.}
All experiments use LLaMA~3.1-8B Instruct \cite{grattafiori2024llama3} and Qwen 3-4B \cite{yang2025qwen3} models {covering complementary model families with different scales and multilingual training distributions, enabling us to test whether observed cross-lingual reasoning behaviors generalize beyond a single architecture. }

\paragraph{Datasets.}
We evaluate on two {parallel multilingual MCQA} datasets: Belebele \cite{bandarkar-etal-2024-belebele} and MMLU-ProX(-Lite) \cite{xuan-etal-2025-mmlu}. Across both datasets, we study three groups of 9 languages: High (English, French, Spanish, Hindi), Mid (Bengali, Afrikaans, Swahili), and Low (Urdu, Yoruba), grouped using LLaMA-3.1-8B \cite{grattafiori2024llama3} as a training-resource proxy as prior work shows that multilingual performance is shaped not only by conventional resource level, but also by pretraining exposure, model size, token similarity, and sociolinguistic factors \cite{nezhad2025beyond}. {This selection ensures diversity not only across resource tiers but also across geographic regions and language families, spanning Indo-European, Indo-Aryan, and Niger–Congo languages. } 

{\noindent\textbf{\em Belebele}~is a machine-translated reading-comprehension datasets that enables controlled translation and resource-tier comparisons.

\noindent\textbf{\em MMLU-ProX(-Lite)} is a multilingual, knowledge-intensive MCQA dataset with 10 answer options} that stresses reasoning under domain knowledge and translation noise.

\paragraph{Answer scoring and decoding.} Both datasets are MCQA with finite digit-based option set. We force decode with ``[['' and score answers only the option digits at the final position. During our preliminary testing, this prevented blank or malformed outputs and avoided accidental prompt repetition, ensuring that predictions are directly comparable across examples. For each prompt (and layer) we compute logit-lens probabilities over the options and apply post-hoc temperature scaling on a held-out calibration split (20\%).

\paragraph{Metrics.} We report Accuracy and ECE, along with other reliability metrics such as confidence and entropy \cite{guo2017calibration}. Accuracy measures correctness of the chosen option whereas ECE measures how well confidence matches empirical accuracy \cite{guo2017calibration, zhou2025beyond}. We compute ECE as the bin-weighted gap between mean accuracy and mean confidence within confidence bins. Confidence is the predicted probability of the selected option.  Entropy summarizes uncertainty over the options \cite{geng-etal-2024-survey,zhang2025all}. Detailed definitions are included in Appendix~\ref{sec:appendix_metrics}.

\begin{figure}[!t]
\centering
\begin{tcolorbox}[mypromptbox]
\small\ttfamily
\textbf{Passage (P):} ``Asynchronous communication encourages time for reflection and reaction to others. It allows students the ability to work at their own pace \dots''\\
\textbf{Question (Q):} ``Which of the following is not a benefit of asynchronous communication for students?''\\
\textbf{Choices (C):} A) The use of internet as a resource \quad
B) Face-to-face access to instructors at any time of day \quad
C) Flexible working hours \quad
D) Pace control
\end{tcolorbox}
\caption{An example from the Belebele dataset illustrating passage, question, and choices.} 
\label{fig:belebele_translation_scopes}
\end{figure}

\section{Analysis: Text scope$\times$Language mode} 
\subsection{Analysis Setup}
Our analysis considers  \emph{Text scope} and \emph{Language mode}.~These two factors are often conflated in multilingual QA, but we vary them independently. {Appendix~\ref{sec:appendix_prompt_templates} illustrates this setup with a worked Belebele example and abridged prompt templates for the T, EN, and TEN modes.}

\begin{itemize}[leftmargin=*, itemsep=2pt]
\item \textbf{Text scope}.  We selectively replace only input text fields with their English translations while keeping the remaining fields in the target language.
For Belebele, we use five settings: Q, P, QP, C, and All (see an example in Figure~\ref{fig:belebele_translation_scopes}).
For MMLU-ProX-Lite which has no passage, we use Q, C, and All.
\item \textbf{Language mode}. We vary the language modes as target-only (T), English-only (EN), or bilingual (TEN) while keeping the instructions in English only {to match the models’ dominant latent space and reduce cross-lingual variability introduced by localized instruction text \cite{wendler-etal-2024-llamas,agrawal2024evaluating,qin-etal-2023-cross}}.
\end{itemize}

\paragraph{Correctness Taxonomy.}~For each instance we record correctness under each translation mode:
$(c_T, c_{EN}, c_{TEN}) \in \{0,1\}^3$, where `1' indicates the model selected the correct answer.~We map these patterns into five categories that explain how language affects success or failure:

\begin{figure*}[!t]
    \centering
    \includegraphics[width=1\linewidth]{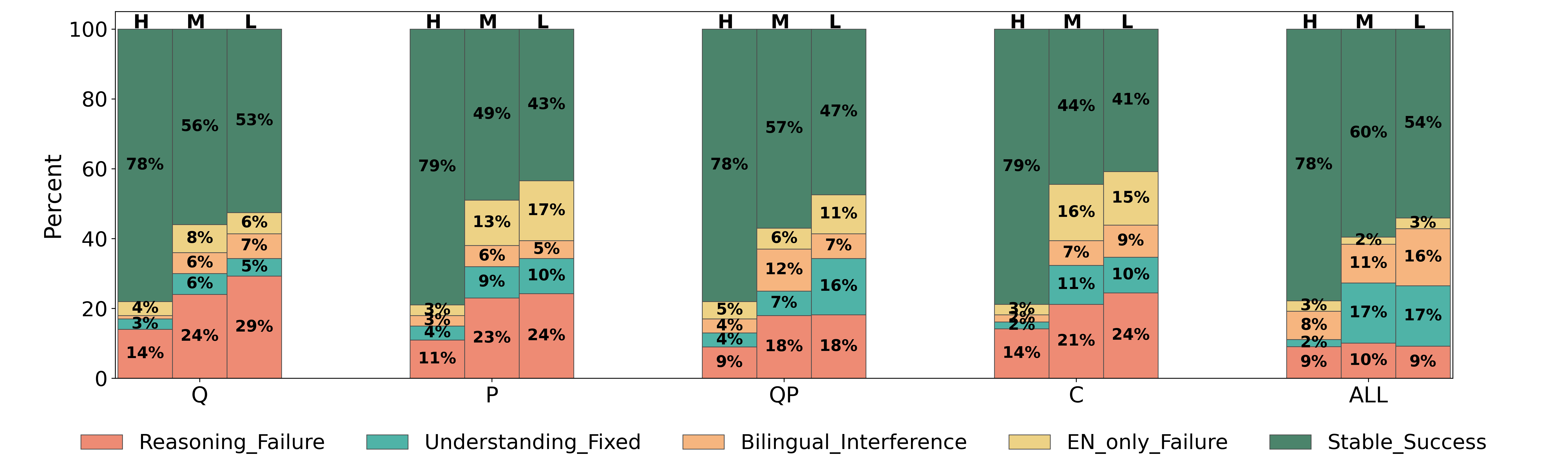}
    \caption{Correctness taxonomy by resource tier (H/M/L) across text scopes [Q/P/QP/C/All] for Belebele dataset using LLama model.}
    \label{fig:root-cause}
\end{figure*}


\begin{table*}[!t]
\centering
\small
\begin{tabular}{l|cc|cc|cc|cc|cc|cc}
\toprule
 & \multicolumn{6}{c|}{\textbf{LLaMA~3.1-8B Instruct}} & \multicolumn{6}{c}{\textbf{Qwen~3-4B}}\\
\cmidrule(lr){2-7}
\cmidrule(lr){8-13}
Resource & \multicolumn{2}{c|}{T} & \multicolumn{2}{c|}{EN} & \multicolumn{2}{c|}{TEN} & \multicolumn{2}{c|}{T} & \multicolumn{2}{c|}{EN} & \multicolumn{2}{c}{TEN} \\
 & Acc $\uparrow$ & ECE $\downarrow$ & Acc$\uparrow$ & ECE $\downarrow$ & Acc$\uparrow$ & ECE $\downarrow$& Acc$\uparrow$ & ECE $\downarrow$& Acc$\uparrow$ & ECE $\downarrow$& Acc$\uparrow$ & ECE $\downarrow$\\
\midrule
\multicolumn{13}{c}{\bf Belebele} \\
\midrule
High & {81.26} & {9.15} & {\textbf{81.79}} & {10.05} & {81.78} & \textbf{8.75} & {80.28} & {9.14} & {87.73} & \textbf{7.43} & {\textbf{88.69}} & {8.25} \\
Mid & \textbf{68.88} & 14.05 & 65.15 & 12.42 & 61.79 & \textbf{9.81} & 66.28 & 10.67 & \textbf{86.55} & \textbf{8.64} & 79.86 & 10.92 \\
Low & \textbf{44.59} & \textbf{12.36} & 44.30 & 12.81 & 44.30 & 14.05 & 45.95 & 9.42 & \textbf{85.44} & \textbf{6.48} & 81.65 & 9.31 \\
\midrule
\multicolumn{13}{c}{\bf MMLU-ProX-Lite} \\
\midrule
High & {34.65} & \textbf{11.02} & 31.02 & 13.19 & {\textbf{38.84}} & 12.16 & {37.78} & 13.83 & {36.27} & \textbf{13.02} & \textbf{45.12} & {15.06} \\
Mid & 25.64 & 13.13 & {\textbf{35.19}} & \textbf{10.44} & 31.57 & {11.67} & 32.67 & \textbf{13.82} & 34.52 & 16.69 & {\textbf{46.38}} & 21.51 \\
Low & 23.47 & \textbf{10.02} & 29.35 & 13.10 & \textbf{33.70} & 13.46 & 28.57 & \textbf{12.81} & 35.87 & 18.24 & \textbf{44.57} & 17.63 \\
\bottomrule
\end{tabular}
\caption{Analysis resource-tier results for Belebele and MMLU-ProX-Lite (LLaMA/Qwen). Accuracy and ECE for High/Mid/Low tiers under T/EN/TEN; bold marks the best accuracy and lowest ECE  per resource tier (row-wise).}
\label{tab:module1_res_main}
\end{table*}

\begin{figure*}[!t]
\centering
\includegraphics[width=1\textwidth, height =10.5cm]{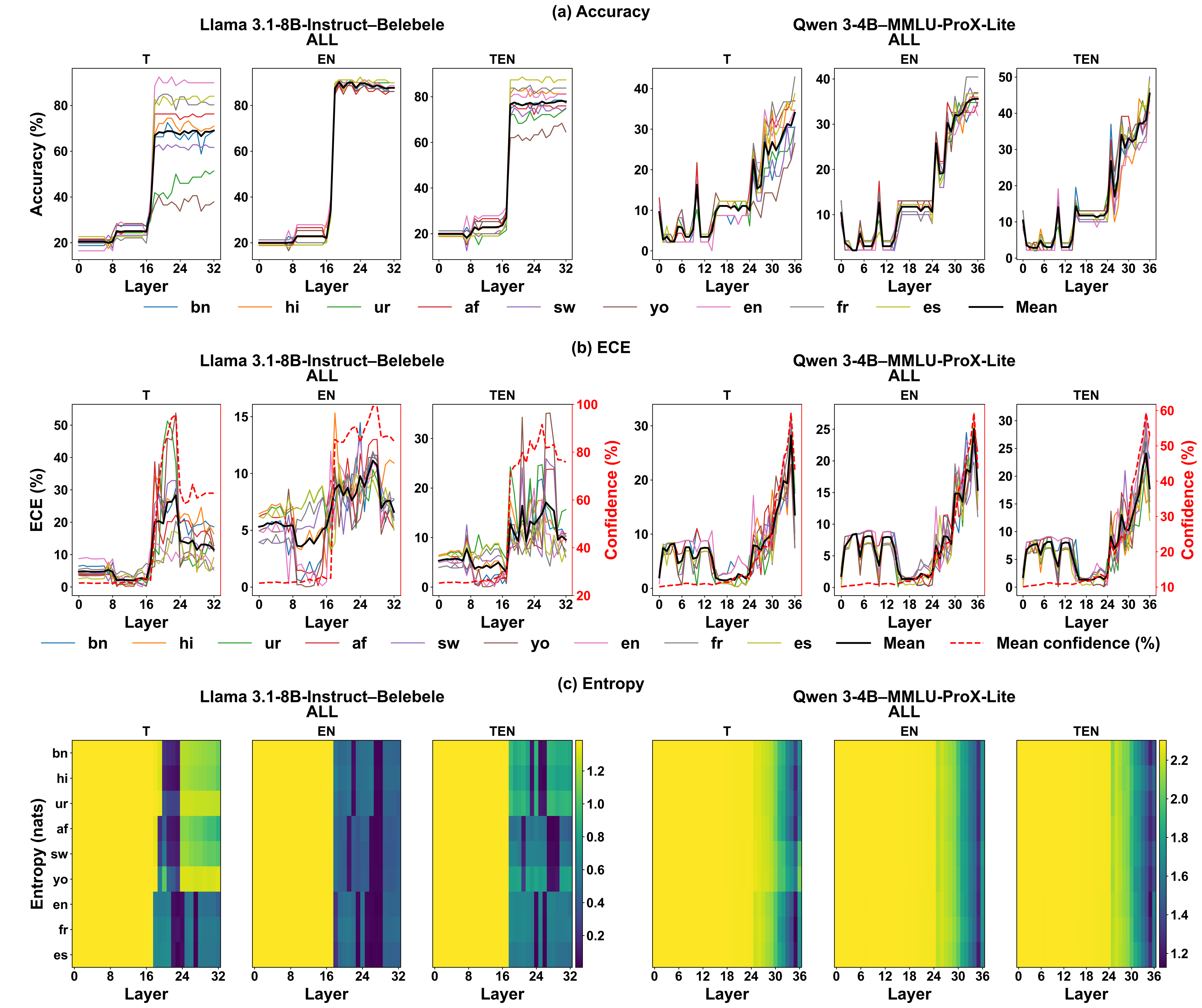}
\caption{Layer‑wise comparison for Qwen‑MMLUProXLite and LLaMA‑Belebele under `ALL' mode : (a)~accuracy curves per dialect with a black mean; (b)~ECE curves per dialect with a black mean and mean confidence overlaid (red dashed); (c)~entropy heatmaps (dialect × layer). All metrics are aggregated over dialects from logit‑lens probes, highlighting late‑layer accuracy gains alongside calibration and uncertainty shifts.}
\label{fig:layer_all_vs_choices}
\end{figure*}

\begin{itemize}[leftmargin=*, itemsep=2pt]
\item \texttt{Reasoning Failure} $(0,0,0)$: the model fails under all language modes indicating an incorrect answer.
\item \texttt{Understanding Fixed} $(0,1,1)$ or $(0,0,1)$: target‑only fails, but English or bilingual prompting succeeds suggesting that translation helps understanding.
\item \texttt{Bilingual Interference} $(1,1,0)$ or $(0,1,0)$: the bilingual prompt hurts relative to a monolingual prompt, thus introducing interference.
\item \texttt{EN only Failure} $(1,0,0)$ or $(1,0,1)$: English prompting fails despite target‑only succeeding.
\item \texttt{Stable Success} $(1,1,1)$: the model succeeds under all language modes indicating that the example is robust to language.
\end{itemize}

We use these categories as behavioral signatures of model performance under different language modes. These are not causal diagnoses, since errors may also arise from translation artifacts or ambiguity. We include the remaining model--dataset root-cause plots in Appendix~\ref{sec:root-cause}.

\subsection{Main Findings}
\label{sec:module1_findings}
We vary text scope and language mode to analyze accuracy and calibration across resource tiers, models, and benchmarks.

\paragraph{Correctness analysis.}
Figure~\ref{fig:root-cause} shows that the effects of translation and structured processing vary by resource level and translation scope. Stable success is highest for high-resource languages and drops for mid- and low-resource ones, reflecting weaker underlying representations. At the same time, translation helps more in lower-resource settings by fixing understanding errors, where language comprehension is often the main bottleneck. Reasoning failures persist across tiers, showing that translation alone does not resolve many errors.

Text scope also changes the type of errors. Translating only the question {is insufficient in evidence-grounded {MCQA task} because it mismatches an English query with non-English evidence, weakening grounding. Broader translation reduces these failures and fixes more understanding errors, but it also increases bilingual interference—especially in low-resource languages—sometimes destabilizing otherwise correct predictions.

\paragraph{Resource-tier effects.}
Table~\ref{tab:module1_res_main} shows that the effect of using \textsc{T}, \textsc{EN}, or \textsc{TEN} depends strongly on the dataset, the model, and the resource tier. There is no single language mode that works best in all cases. On Belebele, LLaMA gains little from translating into English: performance is similar for high-resource languages and does not improve for low-resource ones, although calibration sometimes improves (for example, \textsc{TEN} lowers ECE in the high and mid tiers). Qwen behaves differently: it benefits most from English prompting in mid- and low-resource languages, where \textsc{EN} leads to large accuracy gains and often better calibration. 

On MMLU-ProX-Lite, the pattern changes again. Here, \textsc{TEN} usually gives the best accuracy for both models, consistent with the idea that selectively mixing English support can help on multi-step reasoning questions. However, these  gains often come with worse calibration (higher ECE), showing a clear trade-off between correctness and reliability. Overall, these results suggest that English or bilingual support is not uniformly helpful, and that a fixed inference language mode is likely suboptimal.

\paragraph{Layer- and language-level patterns.} Figure~\ref{fig:layer_all_vs_choices} shows {a consistent depth-wise misalignment between accuracy, sharpness, and calibration. In Figure~\ref{fig:layer_all_vs_choices}a, accuracy generally improves into later layers, although the peak depends on the setting: for LLaMA-Belebele, EN peaks around layer 19, TEN around layer 31, and T only at the final layer, while for Qwen-MMLUProXLite accuracy continues rising through the upper layers. In contrast, Figure~\ref{fig:layer_all_vs_choices}b shows that calibration is often best earlier, with mean ECE minimized around layers 10--17 for LLaMA-Belebele and 17--19 for Qwen-MMLUProXLite, after which confidence rises faster than reliability; across models and datasets, TEN also often yields lower ECE than EN. 

Figure~\ref{fig:layer_all_vs_choices}c shows that entropy continues to decline into the mid-to-upper layers --- roughly 22--26 for LLaMA and 24--30 for Qwen --- so sharper predictions are not necessarily better calibrated. This gap is largest in lower-resource languages, which display noisier layer trajectories and worse final-layer calibration than higher-resource ones. Overall, these results suggest that depth improves prediction sharpness more consistently than reliability, motivating inference policies that prioritize intermediate-layer calibration over final-layer confidence.  Full-resolution metric-specific panels are provided in Appendix~\ref{sec:appendix_layer_curves}.

\section{Reliability-Aware Adaptive Inference (RAAI)}
We introduce RAAI, a training‑free, test‑time framework that adaptively selects inference strategies based on estimated reliability and risk. RAAI consists of two independently deployable components:
{\RaggedRight
\begin{itemize}[leftmargin=1.2em,itemsep=0.3em,topsep=0.2em]
    \item RAAI-Bilingual Gated Fallback~(\textsc{RouteGate}), which selectively routes inputs from target-only inference (\textsc{T}) to selective bilingual inference (\textsc{TEN}) when reliability signals indicate uncertainty or disagreement; and
    \item RAAI-Sequential Reasoning Gate~(\textsc{SeqGate}), which selectively escalates to a more expensive bilingual sequential reasoning prompt only for groups identified as high-risk.
\end{itemize}
}
The goal is to improve accuracy and reliability while avoiding unnecessary bilingual processing or increased compute on cases that are already stable. 

\subsection{RAAI-\textsc{RouteGate}}
\label{sec:module2_raai_fuse}
We define a compute-adaptive fallback policy over prompt variants. For each input $x$, we first run a
target-only prompt (\textsc{T}) and then estimate the reliability of this prediction {using calibrated
probability signals derived from logit-lens layer predictions.} If ECE remains high, we escalate to a bilingual
prompt (\textsc{TEN}); otherwise, we keep the original \textsc{T} prediction. This design aims to improve
reliability (lower ECE) while avoiding the cost and potential interference of bilingual prompting on low-risk cases.

\paragraph{Low-ECE layer ensemble.}
We reserve a small stratified development set with 20 examples from each resource tier and use it only for calibration-oriented control decisions. For each prompt variant, we compute layer-wise ECE on this development set :
\begin{align}
\mathrm{ECE}_{\ell}
&= \sum_{b=1}^{B}\tfrac{|S_b|}{N}\,
\bigl|\operatorname{acc}(S_b)-\operatorname{conf}(S_b)\bigr|,
\label{eq:ece-layer}
\end{align}
where $\ell$ indexes layers, $B$ is the number of calibration bins, $S_b$ is the set of examples in bin $b$, and $N$ is the number of development examples. For example $i$, let $c_i$ be the temperature-scaled probability assigned to its predicted valid option. The bin-level confidence is
\begin{equation}
\operatorname{conf}(S_b)=\frac{1}{|S_b|}\sum_{i\in S_b}c_i.
\end{equation}

We therefore select the $
K$ layers with lowest ECE:
\begin{align}
\mathcal{L}_K
&= \operatorname*{arg\,min}_{\ell}^{(K)} \mathrm{ECE}_{\ell}
\end{align}
and form an ensemble distribution:
\begin{align}
p_{\text{ens}}(y\mid x)
&= \tfrac{1}{K}\sum_{\ell\in\mathcal{L}_K} p_{\ell}(y\mid x).
\end{align}

This ensemble provides a more reliable alternative prediction and serves as a disagreement signal relative to the final layer.


\begin{table*}[!t]
\centering
\small
\setlength{\tabcolsep}{2pt}
\begin{tabular}{l|cc|cc|cc|cc|cc|cc|cc|cc}
\toprule
& \multicolumn{8}{c|}{\textbf{LLaMA~3.1-8B-Instruct}}
& \multicolumn{8}{c}{\textbf{Qwen3-4B}} \\
\cmidrule(lr){2-9}
\cmidrule(lr){10-17}

Tier
& \multicolumn{2}{c|}{T Static}
& \multicolumn{2}{c|}{Best Static}
& \multicolumn{2}{c|}{\textsc{RouteGate}}
& \multicolumn{2}{c|}{\textsc{SeqGate}}
& \multicolumn{2}{c|}{T Static}
& \multicolumn{2}{c|}{Best Static}
& \multicolumn{2}{c|}{\textsc{RouteGate}}
& \multicolumn{2}{c}{\textsc{SeqGate}} \\

& Acc $\uparrow$ & ECE $\downarrow$
& Acc $\uparrow$ & ECE $\downarrow$
& Acc $\uparrow$ & ECE $\downarrow$
& Acc $\uparrow$ & ECE $\downarrow$
& Acc $\uparrow$ & ECE $\downarrow$
& Acc $\uparrow$ & ECE $\downarrow$
& Acc $\uparrow$ & ECE $\downarrow$
& Acc $\uparrow$ & ECE $\downarrow$ \\
\midrule

\multicolumn{17}{c}{\textbf{Belebele}} \\

\midrule
High & 81.26 & 9.15 & 81.79 & 10.05 & \textbf{83.58} & \textbf{5.44} & 82.58 & 6.52 & 80.28 & 9.14 & \textbf{88.69} & 8.25 & 83.77 & \textbf{5.19} & 88.54 & 7.69\\
Mid & 68.88 & 14.05 & 68.88 & 14.05 & \textbf{73.19} & \textbf{6.56} & 67.68 & 10.62 & 66.28 & 10.67 & \textbf{86.55} & \textbf{8.64} & 69.33 & 10.76 & 62.96 & 11.68\\
Low & 44.59 & 12.36 & 44.59 & 12.36 & 55.19 & \textbf{7.71} & \textbf{69.70} & 11.34 & 45.95 & 9.42 & \textbf{85.44} & \textbf{6.48} & 71.14 & 13.19 & 83.64 & 8.35\\
\midrule
\multicolumn{17}{c}{\bf{MMLU-ProX-Lite}} \\
\midrule
High & 34.65 & 11.02 & \textbf{38.84} & 12.16 & 35.92 & \textbf{5.42} & 33.76 & 10.59 & 37.78 & 13.83 & 45.12 & 15.06 & 42.04 & \textbf{8.44} & \textbf{47.13} & 10.41\\
Mid & 25.64 & 13.13 & \textbf{35.19} & 10.44 & 27.95 & \textbf{3.38} & 33.73 & 8.76 & 32.67 & 13.82 & 46.38 & 21.51 & 41.78 & \textbf{11.09} & \textbf{51.64} & 14.04\\
Low & 23.47 & 10.02 & 33.70 & 13.46 & 28.62 & \textbf{4.00} & \textbf{34.12} & 9.16 & 28.57 & 12.81 & 44.57 & 17.63 & 38.04 & \textbf{10.65} & \textbf{51.36} & 13.49\\
\bottomrule
\end{tabular}
\caption{Comparison of T- and Best- Static baseline (from Table\ref{tab:module1_res_main}, along with the proposed RAAI-\textsc{RouteGate} and RAAI-\textsc{SeqGate} by resource tier. Bold marks the best accuracy and lowest ECE per row.}
\label{tab:m2_m3_res_compare}
\end{table*}

\paragraph{TEN fallback.}
Let $\hat{y}_{\text{final}}$ be the final-layer prediction and $\hat{y}_{\text{ens}}$ the ensemble prediction, with confidences $c_{\text{final}}=\max_y p_{\text{final}}(y\mid x)$ and $c_{\text{ens}}=\max_y p_{\text{ens}}(y\mid x)$. We trigger bilingual fallback if the final prediction is low-confidence, or if ensemble confidently disagrees with the final prediction:
\begin{equation}
\begin{split}
\operatorname{trigger}(x)=
&\left(c_{\text{final}}<\tau_{\text{final}}\right)\ \lor \\
&\left(\hat{y}_{\text{final}}\neq\hat{y}_{\text{ens}} \ \land\ c_{\text{ens}}\ge\tau_{\text{ens}}\right).
\end{split}
\label{eq:route-trigger}
\end{equation}

The thresholds $\tau_{\text{final}}$ and $\tau_{\text{ens}}$ are tuned on a development set. If triggered,
we replace the \textsc{T} prediction with a \textsc{TEN} prediction as TEN results were more consistent in lowest calibration error score Table \ref{tab:module1_res_main}. 
\subsection{RAAI-\textsc{SeqGate}}
\label{sec:module3}
Even with selective bilingual fallback, some inputs remain difficult due to deeper reasoning failures and calibration instability, especially in specific language and resource-tier groups. \textsc{SeqGate} targets these residual failures by selectively escalating to a more expensive bilingual sequential reasoning prompt (SEQ) only when the risk is high.


SEQ is a single prompt template with three phases: target-language reasoning, English reasoning over translated content, and a reconciliation step producing a final answer. This provides structured cross-lingual reasoning while preserving consistency across representations (see Appendix \ref{sec:seq_prompt} for more details).

SEQ is substantially more expensive than single-pass inference and can introduce instability when applied indiscriminately. We therefore gate SEQ using a group-level risk index computed offline from target-only (T) language mode, without adding test-time cost.

\paragraph{Risk Index.}
For each resource tier $r$ , we define:
\[\mathrm{RI}_r
= \mathrm{RF}^{(T)}_r \times \mathrm{ECE}^{(T)}_{\max,r}
\times \left(1 - \frac{H^{(T)}_r}{\ln |\mathcal{Y}|}\right).\]
Here $\mathrm{RF}^{(T)}_r$ is the fraction of examples labeled \textsc{Reasoning\_Failure} observed for resource tier $r$ under target-only (T) on a development set, $\mathrm{ECE}^{(T)}_{\max,r}$ is the worst layer-wise calibration error under target-only (T), $H^{(T)}_r$ is the mean predictive entropy under target-only (T), and $|\mathcal{Y}|$ is the number of answer options. 
High values identify groups that are error-prone, poorly calibrated, and confidently wrong. We apply SEQ only for groups exceeding a fixed risk threshold.

\section{Results and Discussion}
We evaluate the two RAAI interventions  as {independent} test-time upgrades. 


\subsection{Main Results of RAAI}
Table~\ref{tab:m2_m3_res_compare} compares the best static language setting (\textsc{T}, \textsc{EN}, or \textsc{TEN} from Table~\ref{tab:module1_res_main}) with the two RAAI components—\textsc{RouteGate} and \textsc{SeqGate}—reporting both accuracy ($\uparrow$) and ECE ($\downarrow$). Their roles are distinct: \textsc{RouteGate} and \textsc{SeqGate} outperform the best static baseline in several settings, with the largest gains appearing in mid- and low-resource settings. 

\textsc{RouteGate} is a per-example online fallback mechanism that uses low-ECE logit-lens confidence and disagreement signals to improve reliability, whereas \textsc{SeqGate} is a per-group offline Risk-Index gate that triggers a more expensive sequential bilingual reasoning prompt only for high-risk slices. This contrast is most pronounced in the lowest-resource settings: on Belebele-Low, \textsc{SeqGate} improves accuracy by +25.1 points for LLaMA and +37.7 for Qwen, while on MMLU-ProX-Lite-Low the gains are +10.6 and +22.8  over T, respectively. Detailed results with full metrics are provided in Appendix~\ref{sec:appendix_resource_family}, and routing rates and confidence intervals, RI sensitivity, and inference-cost analysis are reported in Appendix~\ref{app:robustness}.
\begin{table*}[!t]
\centering
\small
\setlength{\tabcolsep}{4pt}
\begin{tabular}{l|ccccc|ccccc}
\toprule
 & \multicolumn{5}{c|}{\textbf{LLaMA~3.1-8B-Instruct}} & \multicolumn{5}{c}{\textbf{Qwen3-4B}} \\
\cmidrule(lr){2-6}
\cmidrule(lr){7-11}
Tier & $\rho(\mathrm{RI}, \mathrm{Acc}\!\downarrow)$ & $\rho(\mathrm{RI}, \mathrm{ECE}\!\uparrow)$ & N & MedErr & AUROC $\uparrow$& $\rho(\mathrm{RI}, \mathrm{Acc}\!\downarrow)$ & $\rho(\mathrm{RI}, \mathrm{ECE}\!\uparrow)$ & N & MedErr & AUROC$\uparrow$ \\
\midrule
\multicolumn{11}{c}{\bf{Belebele}}\\
\midrule
High & {-0.54 (0.013)} & {+0.55 (0.012)} & 20 & 17.88 & 0.73
     & -0.29 (0.207) & -0.13 (0.583) & 20 & 18.64 & 0.58 \\
Mid  & -0.21 (0.457) & -0.21 (0.457) & 15 & 31.25 & 0.50
     & {-0.55 (0.033)} & {+0.75 (0.001)} & 15 & 30.00 & 0.60 \\
Low  & \textbf{-0.58 (0.275) }& \textbf{+0.68 (0.275)} & 10 & 55.41 & 0.72
     & \textbf{-0.87 (0.001)} & \textbf{+0.74 (0.014)} & 10 & 54.05 & 1.00 \\
\midrule
\multicolumn{11}{c}{\bf{MMLU-ProX-Lite}}\\
\midrule
High & -0.19 (0.544) & -0.11 (0.738) & 12 & 66.33 & 0.53
     & {-0.78 (0.003)} & {+0.78 (0.003)} & 12 & 63.22 & 1.00 \\
Mid  & +0.05 (0.893) & {-0.74 (0.023)} & 9 & 73.91 & 0.22
     & -0.05 (0.893) & \textbf{-0.69 (0.042)} & 9 & 69.39 & 0.78 \\
Low  & \textbf{-0.88 (0.021)} & \textbf{+0.80 (0.021)} & 6 & 76.53 & 1.00
     & \textbf{-0.84 (0.021)} & \textbf{+0.88 (0.02)} & 6 & 71.43 & 1.00 \\
\bottomrule
\end{tabular}
\caption{RI validation by resource tier: Spearman $\rho$ with Accuracy (Acc) and ECE, N, Median Error (MedErr), and AUROC (p-values in parentheses) for LLaMA and Qwen. Median Error and AUROC use within each resource tier as the positive-class threshold.}
\label{tab:RI_corr_spearman_resource}
\end{table*}

\paragraph{\textsc{RouteGate} improves calibration} Across both models and datasets, \textsc{RouteGate} consistently reduces ECE relative to the \textsc{T} baseline. On Belebele, \textsc{RouteGate} lowers ECE in all tiers {even by ~7.5 points on mid-resource}; on MMLU, the effect is even more pronounced {(ECE drops by ~10 points in mid-resource)}, suggesting that bilingual fallback stabilizes the model's confidence, making predicted probabilities better aligned with actual correctness. Accuracy also often improves, but more modestly and less uniformly than calibration. 

\paragraph{\textsc{SeqGate} primarily improves accuracy, especially in low-resource tiers} \textsc{SeqGate} delivers the largest accuracy gains, particularly in mid- and low-resource settings. \textsc{SeqGate} is never applied uniformly; it is only triggered on inputs that the Risk Index flags as high-risk. Across datasets, it improves low-resource tiers scores by 10\% (Llama) and 22\% (Qwen) over T. These gains indicate that sequential bilingual reasoning is especially helpful when baseline performance is weakest, likely because these cases require deep reasoning and benefit from additional structured processing.

\paragraph{Accuracy-calibration trade-off between \textsc{RouteGate} and \textsc{SeqGate}} A clear trade-off emerges: {\textsc{RouteGate}} achieves the lowest ECE in most rows but smaller accuracy gains. {\textsc{SeqGate}} achieves the highest accuracy in most rows but often with higher ECE than \textsc{RouteGate}. 
\paragraph{Resource tiers} Both methods show larger gains in mid- and low-resource tiers. High-resource tiers often already perform well under \textsc{T}, leaving less room for improvement. Low-resource tiers benefit the most from both bilingual support and structured reasoning.
\paragraph{Model differences} Qwen starts from higher baseline accuracy and calibration in most settings, so relative improvements are smaller than on LLaMA. However, \textsc{SeqGate} still produces large absolute gains for Qwen in low-resource tiers, indicating that even strong multilingual models struggle on hard low-resource reasoning tasks.

\subsection{\textsc{SeqGate} Risk Index Validation} 
\label{sec:results_mod3}
\textsc{SeqGate} focuses on residual failures that persist even after English-only or bilingual mode where \textsc{T}, \textsc{EN}, and \textsc{TEN} all break down—primarily in low-resource scenarios. 

For each language group, we compute RI from the target-only (T) development logs. We save the gold label, predicted option, option probabilities, confidence, and entropy. We first computed the reasoning failure from layer probing, along with calculating ten maximum ECE values across layers and high entropy to find all risk components . We then freeze the group-level Risk-Index(RI) score.

Table~\ref{tab:RI_corr_spearman_resource} supports using RI as a \emph{gating} policy rather than deploying {\textsc{SeqGate} across the board: in the hardest Low-resource tier slices for both models, RI attains perfect discrimination (AUROC=$1.00$), exactly where \textsc{SeqGate} yields its largest improvements . However, \textsc{Seq} is not uniformly advantageous and can degrade calibration compared to the \textsc{RouteGate}, underscoring the need for risk-based escalation.

To validate further, RI as a robust score from \textsc{T}-only statistics,  we report correlations with \textsc{T} accuracy/ECE and AUROC for separating higher- vs.\ lower-error groups in Table~\ref{tab:RI_corr_spearman_resource}. RI is consistently negative with accuracy across models, while its link to ECE is benchmark-dependent (stronger on Belebele; weaker or sometimes reversed on MMLU-ProX-Lite), suggesting dataset-specific calibration pathologies. AUROC remains high; tier-pooled estimates are noisier for small $N$ but preserve the same directional signal.Appendix~\ref{app:ri-robustness} reports the factor ablation and routing-budget sensitivity used to assess this composite score.

\begin{table}[!t]
\centering
\small
\begin{tabularx}{\linewidth}{p{0.05\linewidth} X}
\toprule
\multicolumn{2}{l}{\textbf{Example 1 (Bengali): T works, EN fails}} \\
\midrule
(P) &
Vision, or the ability to see depends on visual system sensory organs or eyes … \\
(Q) &
What is the ability to see dependent upon? \\
(C) &
A) Visual system sensory organs \;
B) The requirements of an organism \;
C) A sensitivity to certain wave-lengths \;
D) Varying degrees of acuity \\
Ans. &
A \\
Pred. &
T = A (correct); EN = blank/invalid; TEN = A (correct). \\
\midrule
\multicolumn{2}{l}{\textbf{Example 2 (Yoruba): T fails, EN fails, TEN works}} \\
\midrule
(P) &
The story presented in the French opera … the performers smoke cannabis joints on stage … \\
(Q) &
What do performers encourage the audience to do during Camille Saint-Saens’ opera? \\
(C) &
A) Partake in the use of cannabis \;
B) Take a trip to Japan \;
C) Join them onstage for the performance \;
D) Allow their lives to be dictated by what they love \\
Ans. &
A \\
Pred. &
T = blank/invalid; EN = blank/invalid; TEN = A (correct). \\
\bottomrule
\end{tabularx}
\caption{Qualitative analysis showing example instances. Examples shown in English.}
\label{tab:example_instances}
\end{table}

\subsection{Qualitative Analysis} 
Table~\ref{tab:example_instances} shows why fixed prompting is unreliable. In the Bengali example, target-only succeeds while English-only fails, showing that added English can destabilize a correct prediction. In the Yoruba example, both target-only and English-only fail, but bilingual prompting succeeds, suggesting that TEN can better preserve grounding in some cases. Together, these examples reinforce our main finding: multilingual reasoning benefits not from more English everywhere, but from selective intervention guided by reliability and risk.

\section{Conclusion}
\label{sec:discussion}

Our work provides a diagnostic-to-deployment framework for multilingual reasoning: controlled analysis shows that English-centric assistance can help, but its gains are setting-dependent and often introduce interference or calibration costs, making fixed prompting unreliable. We therefore propose \textsc{RAAI}, a training-free test-time framework that adaptively allocates language assistance and computation based on risk. \textsc{RouteGate} is a low-cost per-example controller for improving calibration through selective bilingual fallback, while \textsc{SeqGate} is a higher-cost per-group controller that reserves sequential bilingual reasoning for the hardest slices, improving accuracy especially in low-resource settings. Together, they show that robust multilingual reasoning is better achieved through adaptive intervention than through any single fixed prompting strategy. This suggests that multilingual reasoning is best supported not by more assistance everywhere, but by using the right assistance only where it is needed most.

\section*{Limitations}
Our controlled experiments use benchmark-provided parallel target-language and English fields. Consequently, the reported cost analysis covers LLM inference but excludes the latency and expense of producing translations. A deployed EN, TEN, or SEQ system would require cached translations or an external translation service, and translation quality could affect both predictions and routing decisions \cite{agrawal2023all}. Comparing human translations, machine translations, and controlled translation perturbations remains an important extension.

The evaluation is restricted to closed-set multilingual MCQA. This setting provides a shared finite answer space and supports direct option-level confidence, entropy, and ECE measurement, but the findings may not transfer directly to open-ended generation, where semantic equivalence, output language, fluency, and sequence-level confidence introduce additional uncertainty. We also evaluate one model size from each of two families and nine languages. The results therefore provide cross-family replication, not a within-family scaling study or broad language-level coverage. In particular, the Low tier contains only Urdu and Yoruba. Finally, \textsc{SeqGate} uses a development-derived language--scope risk profile; deploying it on a new language or domain requires a small profiling set.

\section*{Ethics Statement}
We analyze multilingual behavior to improve equity across languages and dialects, particularly for low-resource and historically under-served communities.
Our work is purely observational and inference-time: we do not collect new user data or train models on sensitive content.
Nonetheless, better multilingual performance can have downstream impacts in high-stakes domains; we therefore report calibration and risk measures, not only accuracy, and we will release code and configurations to support responsible use. Reproducibility and tool-use details are provided in Appendix~\ref{app:reproducibility}.

\section*{Acknowledgments}
We sincerely thank Rhitabrat Pokharel for his valuable contributions and insightful discussions. We also thank the reviewers for their helpful feedback and gratefully acknowledge partial support from the Wedge Vision Award.

\bibliography{latex/custom}

@article{zhou2025beyond,
  title = "Beyond the Final Layer: Intermediate Representations for Better Multilingual Calibration in Large Language Models",
  author = "Zhou, Ej and Zhang, Caiqi and Hu, Tiancheng and Li, Chengzu and Collier, Nigel and Vuli{\'c}, Ivan and Korhonen, Anna",
  journal = "arXiv preprint arXiv:2510.03136",
  year = "2025",
  doi = "10.48550/arXiv.2510.03136",
  url = "https://arxiv.org/abs/2510.03136"
}

@inproceedings{bandarkar2024belebele,
  title     = {The Belebele Benchmark: a Parallel Reading Comprehension Dataset in 122 Language Variants},
  author    = {Bandarkar, Lucas and Liang, Davis and Muller, Benjamin and Artetxe, Mikel and Shukla, Satya Narayan and Husa, Donald and Goyal, Naman and Krishnan, Abhinandan and Zettlemoyer, Luke and Khabsa, Madian},
  booktitle = {Proceedings of the 62nd Annual Meeting of the Association for Computational Linguistics (Volume 1: Long Papers)},
  pages     = {749--775},
  year      = {2024},
  address   = {Bangkok, Thailand},
  publisher = {Association for Computational Linguistics},
  doi       = {10.18653/v1/2024.acl-long.44},
  url       = {https://aclanthology.org/2024.acl-long.44}
}

@article{schut2025multilingual,
  title   = {How Do Large Language Models Handle Multilingualism?},
  author  = {Zhao, Yiran and Zhang, Wenxuan and Chen, Guizhen and Kawaguchi, Kenji and Bing, Lidong},
  journal = {Advances in Neural Information Processing Systems},
  year    = {2024},
  note    = {NeurIPS 2024},
  url     = {https://arxiv.org/abs/2402.18815}
}

@inproceedings{geng-etal-2024-survey,
    title = "A Survey of Confidence Estimation and Calibration in Large Language Models",
    author = "Geng, Jiahui  and
      Cai, Fengyu  and
      Wang, Yuxia  and
      Koeppl, Heinz  and
      Nakov, Preslav  and
      Gurevych, Iryna",
    editor = "Duh, Kevin  and
      Gomez, Helena  and
      Bethard, Steven",
    booktitle = "Proceedings of the 2024 Conference of the North American Chapter of the Association for Computational Linguistics: Human Language Technologies (Volume 1: Long Papers)",
    month = jun,
    year = "2024",
    address = "Mexico City, Mexico",
    publisher = "Association for Computational Linguistics",
    url = "https://aclanthology.org/2024.naacl-long.366/",
    doi = "10.18653/v1/2024.naacl-long.366",
    pages = "6577--6595"
}

@article{yang2025qwen3,
  title={Qwen3 technical report},
  author={Yang, An and Li, Anfeng and Yang, Baosong and Zhang, Beichen and Hui, Binyuan and Zheng, Bo and Yu, Bowen and Gao, Chang and Huang, Chengen and Lv, Chenxu and others},
  journal={arXiv preprint arXiv:2505.09388},
  year={2025}
}

@article{kang2025gaps,
  title   = {Why Do Multilingual Reasoning Gaps Emerge in Reasoning Language Models?},
  author  = {Kang, Deokhyung and Hwang, Seonjeong and Kim, Daehui and Kim, Hyounghun and Lee, Gary Geunbae},
  journal = {arXiv preprint arXiv:2510.27269},
  year    = {2025},
  doi     = {10.48550/arXiv.2510.27269},
  url     = {https://arxiv.org/abs/2510.27269}
}

@inproceedings{zhang2025all,
  title={All Roads Lead to Rome: Graph-Based Confidence Estimation for Large Language Model Reasoning},
  author={Zhang, Caiqi and Shu, Chang and Shareghi, Ehsan and Collier, Nigel},
  booktitle={Proceedings of the 2025 Conference on Empirical Methods in Natural Language Processing},
  pages={31802--31812},
  year={2025}
}

@inproceedings{qin-etal-2023-cross,
  title     = {Cross-lingual Prompting: Improving Zero-shot Chain-of-Thought Reasoning across Languages},
  author    = {Qin, Libo and Chen, Qiguang and Wei, Fuxuan and Huang, Shijue and Che, Wanxiang},
  booktitle = {Proceedings of the 2023 Conference on Empirical Methods in Natural Language Processing},
  pages     = {2695--2709},
  year      = {2023},
  address   = {Singapore},
  publisher = {Association for Computational Linguistics},
  doi       = {10.18653/v1/2023.emnlp-main.163},
  url       = {https://aclanthology.org/2023.emnlp-main.163}
}

@inproceedings{zhu-etal-2024-question,
  title     = {Question Translation Training for Better Multilingual Reasoning},
  author    = {Zhu, Wenhao and Huang, Shujian and Yuan, Fei and She, Shuaijie and Chen, Jiajun and Birch, Alexandra},
  booktitle = {Findings of the Association for Computational Linguistics: ACL 2024},
  pages     = {8411--8423},
  year      = {2024},
  address   = {Bangkok, Thailand},
  publisher = {Association for Computational Linguistics},
  doi       = {10.18653/v1/2024.findings-acl.498},
  url       = {https://aclanthology.org/2024.findings-acl.498}
}

@inproceedings{bandarkar2024layer,
  title     = {Layer Swapping for Zero-Shot Cross-Lingual Transfer in Large Language Models},
  author    = {Bandarkar, Lucas and Muller, Benjamin and Yuvraj, Pritish and Hou, Rui and Singhal, Nayan and Lv, Hongjiang and Liu, Bing},
  booktitle = {Proceedings of the 13th International Conference on Learning Representations},
  year      = {2025},
  note      = {To appear; preprint arXiv:2410.01335},
  url       = {https://arxiv.org/abs/2410.01335}
}

@inproceedings{yang2023multilingual,
  title     = {On the Calibration of Multilingual Question Answering {LLMs}},
  author    = {Yang, Yahan and Dan, Soham and Roth, Dan and Lee, Insup},
  booktitle = {arXiv preprint arXiv:2311.08669},
  year      = {2023},
  url       = {https://arxiv.org/abs/2311.08669}
}

@inproceedings{ahuja-etal-2022-calibration,
  title     = {On the Calibration of Massively Multilingual Language Models},
  author    = {Ahuja, Kabir and Sitaram, Sunayana and Dandapat, Sandipan and Choudhury, Monojit},
  booktitle = {Proceedings of the 2022 Conference on Empirical Methods in Natural Language Processing},
  pages     = {4310--4323},
  year      = {2022},
  address   = {Abu Dhabi, United Arab Emirates},
  publisher = {Association for Computational Linguistics},
  doi       = {10.18653/v1/2022.emnlp-main.290},
  url       = {https://aclanthology.org/2022.emnlp-main.290}
}

@article{xue2024calib,
  title   = {A Comprehensive Study of Multilingual Confidence Estimation on Large Language Models},
  author  = {Xue, Boyang and Wang, Hongru and Wang, Rui and Wang, Sheng and Wang, Zezhong and Du, Yiming and Liang, Bin and Wong, Kam-Fai},
  journal = {arXiv preprint arXiv:2402.13606},
  year    = {2024},
  url     = {https://arxiv.org/abs/2402.13606}
}

@inproceedings{kojima-etal-2024-multilingual,
  title     = {On the Multilingual Ability of Decoder-based Pre-trained Language Models: Finding and Controlling Language-Specific Neurons},
  author    = {Kojima, Takeshi and Okimura, Itsuki and Iwasawa, Yusuke and Yanaka, Hitomi and Matsuo, Yutaka},
  booktitle = {Proceedings of the 2024 Conference of the North American Chapter of the Association for Computational Linguistics: Human Language Technologies (Volume 1: Long Papers)},
  pages     = {6919--6971},
  year      = {2024},
  address   = {Mexico City, Mexico},
  publisher = {Association for Computational Linguistics},
  doi       = {10.18653/v1/2024.naacl-long.384},
  url       = {https://aclanthology.org/2024.naacl-long.384}
}

@article{qi2025reason,
  title   = {When Models Reason in Your Language: Controlling Thinking Trace Language Comes at the Cost of Accuracy},
  author  = {Qi, Jirui and Chen, Shan and Xiong, Zidi and Fern{\'a}ndez, Raquel and Bitterman, Danielle S. and Bisazza, Arianna},
  journal = {arXiv preprint arXiv:2505.22888},
  year    = {2025},
  doi     = {10.48550/arXiv.2505.22888},
  url     = {https://arxiv.org/abs/2505.22888}
}

@inproceedings{guo2017calibration,
  title     = {On Calibration of Modern Neural Networks},
  author    = {Guo, Chuan and Pleiss, Geoff and Sun, Yu and Weinberger, Kilian Q.},
  booktitle = {Proceedings of the 34th International Conference on Machine Learning (ICML)},
  year      = {2017}
}

@inproceedings{xuan-etal-2025-mmlu,
    title = "{MMLU}-{P}ro{X}: A Multilingual Benchmark for Advanced Large Language Model Evaluation",
    author = "Xuan, Weihao  and
      Yang, Rui  and
      Qi, Heli  and
      Zeng, Qingcheng  and
      Xiao, Yunze  and
      Feng, Aosong  and
      Liu, Dairui  and
      Xing, Yun  and
      Wang, Junjue  and
      Gao, Fan  and
      Lu, Jinghui  and
      Jiang, Yuang  and
      Li, Huitao  and
      Li, Xin  and
      Yu, Kunyu  and
      Dong, Ruihai  and
      Gu, Shangding  and
      Li, Yuekang  and
      Xie, Xiaofei  and
      Juefei-Xu, Felix  and
      Khomh, Foutse  and
      Yoshie, Osamu  and
      Chen, Qingyu  and
      Teodoro, Douglas  and
      Liu, Nan  and
      Goebel, Randy  and
      Ma, Lei  and
      Marrese-Taylor, Edison  and
      Lu, Shijian  and
      Iwasawa, Yusuke  and
      Matsuo, Yutaka  and
      Li, Irene",
    editor = "Christodoulopoulos, Christos  and
      Chakraborty, Tanmoy  and
      Rose, Carolyn  and
      Peng, Violet",
    booktitle = "Proceedings of the 2025 Conference on Empirical Methods in Natural Language Processing",
    month = nov,
    year = "2025",
    address = "Suzhou, China",
    publisher = "Association for Computational Linguistics",
    url = "https://aclanthology.org/2025.emnlp-main.79/",
    doi = "10.18653/v1/2025.emnlp-main.79",
    pages = "1513--1532",
    ISBN = "979-8-89176-332-6"
}

@article{kirichenko2025abstentionbench,
  title={AbstentionBench: Reasoning LLMs Fail on Unanswerable Questions},
  author={Kirichenko, Polina and Ibrahim, Mark and Chaudhuri, Kamalika and Bell, Samuel J},
  journal={arXiv preprint arXiv:2506.09038},
  year={2025}
}

@article{tomani2024uncertainty,
  title={Uncertainty-based abstention in llms improves safety and reduces hallucinations},
  author={Tomani, Christian and Chaudhuri, Kamalika and Evtimov, Ivan and Cremers, Daniel and Ibrahim, Mark},
  journal={arXiv preprint arXiv:2404.10960},
  year={2024}
}

@inproceedings{feng2024multilingual,
  title     = {Teaching {LLM}s to Abstain across Languages via Multilingual Feedback},
  author    = {Feng, Shangbin and Shi, Weijia and Wang, Yike and Ding, Wenxuan and Ahia, Orevaoghene and Li, Shuyue Stella and Balachandran, Vidhisha and Sitaram, Sunayana and Tsvetkov, Yulia},
  booktitle = {Proceedings of the 2024 Conference on Empirical Methods in Natural Language Processing},
  month     = nov,
  year      = {2024},
  address   = {Miami, Florida, USA},
  publisher = {Association for Computational Linguistics},
  url       = {https://aclanthology.org/2024.emnlp-main.239/},
  doi       = {10.18653/v1/2024.emnlp-main.239},
  pages     = {4125--4150}
}

@inproceedings{huang-etal-2023-languages,
  title     = "Not All Languages Are Created Equal in {LLM}s: Improving Multilingual Capability by Cross-Lingual-Thought Prompting",
  author    = "Huang, Haoyang  and
               Tang, Tianyi  and
               Zhang, Dongdong  and
               Zhao, Xin  and
               Song, Ting  and
               Xia, Yan  and
               Wei, Furu",
  editor    = "Bouamor, Houda  and
               Pino, Juan  and
               Bali, Kalika",
  booktitle = "Findings of the Association for Computational Linguistics: EMNLP 2023",
  month     = dec,
  year      = "2023",
  address   = "Singapore",
  publisher = "Association for Computational Linguistics",
  url       = "https://aclanthology.org/2023.findings-emnlp.826/",
  doi       = "10.18653/v1/2023.findings-emnlp.826",
  pages     = "12365--12394"
}

@article{grattafiori2024llama3,
  title   = {The Llama 3 Herd of Models},
  author  = {Grattafiori, Aaron and Dubey, Abhimanyu and Jauhri, Abhinav and Pandey, Abhinav and Kadian, Abhishek and Al-Dahle, Ahmad and Letman, Aiesha and others},
  journal = {arXiv preprint arXiv:2407.21783},
  year    = {2024}
}

@article{tam2025chain,
  title   = {Language Matters: How Do Multilingual Input and Reasoning Paths Affect Large Reasoning Models?},
  author  = {Tam, Zhi Rui and Wu, Cheng-Kuang and Chiu, Yu Ying and Lin, Chieh-Yen and Chen, Yun-Nung and Lee, Hung-yi},
  journal = {arXiv preprint arXiv:2505.17407},
  year    = {2025},
  doi     = {10.48550/arXiv.2505.17407},
  url     = {https://arxiv.org/abs/2505.17407}
}

@article{lim2025latent,
  title = "Language-Specific Latent Process Hinders Cross-Lingual Performance",
  author = "Lim, Zheng Wei and Aji, Alham Fikri and Cohn, Trevor",
  journal = "arXiv preprint arXiv:2505.13141",
  year = "2025",
  doi = "10.48550/arXiv.2505.13141",
  url = "https://arxiv.org/abs/2505.13141"
}

@inproceedings{bandarkar-etal-2024-belebele,
  title = "The Belebele Benchmark: a Parallel Reading Comprehension Dataset in 122 Language Variants",
  author = "Bandarkar, Lucas and Liang, Davis and Muller, Benjamin and Artetxe, Mikel and Shukla, Satya Narayan and Husa, Donald and Goyal, Naman and Krishnan, Abhinandan and Zettlemoyer, Luke and Khabsa, Madian",
  editor = "Ku, Lun-Wei and Martins, Andre and Srikumar, Vivek",
  booktitle = "Proceedings of the 62nd Annual Meeting of the Association for Computational Linguistics (Volume 1: Long Papers)",
  month = aug,
  year = "2024",
  address = "Bangkok, Thailand",
  publisher = "Association for Computational Linguistics",
  doi = "10.18653/v1/2024.acl-long.44",
  url = "https://aclanthology.org/2024.acl-long.44",
  pages = "749--775"
}

@inproceedings{shi2023cot,
  title = "Language Models are Multilingual Chain-of-Thought Reasoners",
  author = "Shi, Freda and Suzgun, Mirac and Freitag, Markus and Wang, Xuezhi and Srivats, Suraj and Vosoughi, Soroush and Chung, Hyung Won and Tay, Yi and Ruder, Sebastian and Zhou, Denny and Das, Dipanjan and Wei, Jason",
  booktitle = "Proceedings of the Eleventh International Conference on Learning Representations (ICLR 2023)",
  month = may,
  year = "2023",
  address = "Kigali, Rwanda",
  publisher = "OpenReview.net",
  url = "https://openreview.net/forum?id=fR3wGCk-IXp"
}

@inproceedings{wendler-etal-2024-llamas,
  title = "Do Llamas Work in {E}nglish? On the Latent Language of Multilingual Transformers",
  author = "Wendler, Chris and Veselovsky, Veniamin and Monea, Giovanni and West, Robert",
  editor = "Ku, Lun-Wei and Martins, Andre and Srikumar, Vivek",
  booktitle = "Proceedings of the 62nd Annual Meeting of the Association for Computational Linguistics (Volume 1: Long Papers)",
  month = aug,
  year = "2024",
  address = "Bangkok, Thailand",
  publisher = "Association for Computational Linguistics",
  doi = "10.18653/v1/2024.acl-long.820",
  url = "https://aclanthology.org/2024.acl-long.820",
  pages = "15366--15394"
}

@inproceedings{agrawal2024evaluating,
  title={Evaluating multilingual long-context models for retrieval and reasoning},
  author={Agrawal, Ameeta and Dang, Andy and Nezhad, Sina Bagheri and Pokharel, Rhitabrat and Scheinberg, Russell},
  booktitle={Proceedings of the Fourth workshop on multilingual representation learning (MRL 2024)},
  pages={216--231},
  year={2024}
}

@article{pokharel2026cross,
  title={Cross-lingual activation steering for multilingual language models},
  author={Pokharel, Rhitabrat and Agrawal, Ameeta and Nagar, Tanay},
  journal={arXiv preprint arXiv:2601.16390},
  year={2026}
}

@inproceedings{nezhad2025beyond,
  title={Beyond data quantity: Key factors driving performance in multilingual language models},
  author={Nezhad, Sina Bagheri and Agrawal, Ameeta and Pokharel, Rhitabrat},
  booktitle={Proceedings of the First workshop on language models for low-resource languages},
  pages={225--239},
  year={2025}
}

@inproceedings{pokharel2025capo,
  title={Capo: Confidence aware preference optimization learning for multilingual preferences},
  author={Pokharel, Rhitabrat and Tao, Yufei and Agrawal, Ameeta},
  booktitle={Proceedings of the 14th International Joint Conference on Natural Language Processing and the 4th Conference of the Asia-Pacific Chapter of the Association for Computational Linguistics},
  pages={1144--1156},
  year={2025}
}

@inproceedings{agrawal2023all,
  title={All translation tools are not equal: Investigating the quality of language translation for forced migration},
  author={Agrawal, Ameeta and Singh, Lisa and Jacobs, Elizabeth and Liu, Yaguang and Dunlevy, Gwyneth and Pokharel, Rhitabrat and Uppala, Varun},
  booktitle={2023 IEEE 10th international conference on data science and advanced analytics (DSAA)},
  pages={1--10},
  year={2023},
  organization={IEEE}
}

\appendix

\section{Background: Experimental Protocol, Metrics, and Prompts}
\label{app:protocol}


\subsection{Metric Definitions}
\label{sec:appendix_metrics}
\paragraph{Accuracy.} The fraction of questions the model answers correctly (1 for correct, 0 for incorrect), averaged and scaled by 100.
\[
\mathrm{Acc}\% = 100\cdot \frac{1}{N}\sum_{i=1}^N \mathbf{1}[\hat{y}_i=y_i].
\]
\paragraph{Confidence.}For each question, we take the predicted probability of the chosen option (after temperature scaling and renormalizing over available options), then average and scale by 100.
\[
\begin{aligned}
p_{ij} &= \frac{\exp(l_{ij}/T)}{\sum_{k\in S_i}\exp(l_{ik}/T)}, \\
\hat{y}_i &= \arg\max_{j\in S_i} p_{ij}, \\
c_i &= \max_{j\in S_i} p_{ij}, \\
\mathrm{Conf}\% &= 100\cdot \frac{1}{N}\sum_{i=1}^N c_i .
\end{aligned}
\]
\paragraph{Entropy.} The average uncertainty of the option distribution for each question, computed as Predictive entropy $H(p) = -\sum_{k=1}^{K} p_k\log p_k$ with $K{=}4$ for Belebele.
\paragraph{ECE.} A calibration error that compares average confidence to average accuracy within $M$ uniform confidence bins.
\[
\begin{aligned}
B_m &= \{i \mid c_i\in((m-1)/M,\, m/M]\}, \\
\operatorname{acc}(B_m) &= \frac{1}{|B_m|}\sum_{i\in B_m}\mathbf{1}[\hat{y}_i=y_i], \\
\operatorname{conf}(B_m) &= \frac{1}{|B_m|}\sum_{i\in B_m} c_i, \\
\mathrm{ECE} &= 100\cdot \sum_{m=1}^{M}\frac{|B_m|}{N}
\left|\operatorname{acc}(B_m)-\operatorname{conf}(B_m)\right|.
\end{aligned}
\]
\paragraph{Temperature scaling.} We fit a single scalar temperature on a held-out calibration split (20\%) by minimizing NLL and apply it consistently across modules for confidence/ECE reporting.
\subsection{Example: Scope x Translation Mode (T/EN/TEN)}
\label{sec:appendix_prompt_templates}
Figure~\ref{fig:belebele_translation_scopes_appendix} illustrates the five Belebele translation scopes, and Figure~\ref{fig:prompt_templates_appendix} summarizes the T, EN, and TEN prompt modes.
\begin{figure}[!t]
\small\ttfamily
\centering
\begin{tcolorbox}[title=Scope ]
Passage\_T (fra\_Latn): ``La communication asynchrone favorise le temps de r\'eflexion et de r\'eaction aux autres. Ceci permet aux \'etudiants de travailler \`a leur rythme \dots''\\
Question\_T: ``Lequel des \'el\'ements ci-dessous n'est pas un avantage li\'e \`a la communication asynchrone pour les \'etudiants ?''\\
Choices\_T: A) L'utilisation d'Internet en tant que ressource \quad
B) Des r\'eunions en personne avec les instructeurs \`a tout moment \quad
C) Des horaires de travail flexibles \quad
D) Le travail \`a leur rythme\\
\smallskip
Passage\_EN: ``Asynchronous communication encourages time for reflection and reaction to others. It allows students the ability to work at their own pace \dots''\\
Question\_EN: ``Which of the following is not a benefit of asynchronous communication for students?''\\
Choices\_EN: A) The use of internet as a resource \quad
B) Face-to-face access to instructors at any time of day \quad
C) Flexible working hours \quad
D) Pace control\\
\smallskip
Q: swap in Question\_EN only. \;
P: swap in Passage\_EN only. \;
QP: swap in Passage\_EN + Question\_EN.\\
C: swap in Choices\_EN only. \;
All: swap in Passage\_EN + Question\_EN + Choices\_EN.
\end{tcolorbox}
\caption{Belebele example illustrating translation granularity (Q/P/QP/C/All).}
\label{fig:belebele_translation_scopes_appendix}
\end{figure}

We designate this translation mode template as the T/EN/TEN" scope example set employed in the Belebele dataset.

\begin{figure}[!t]
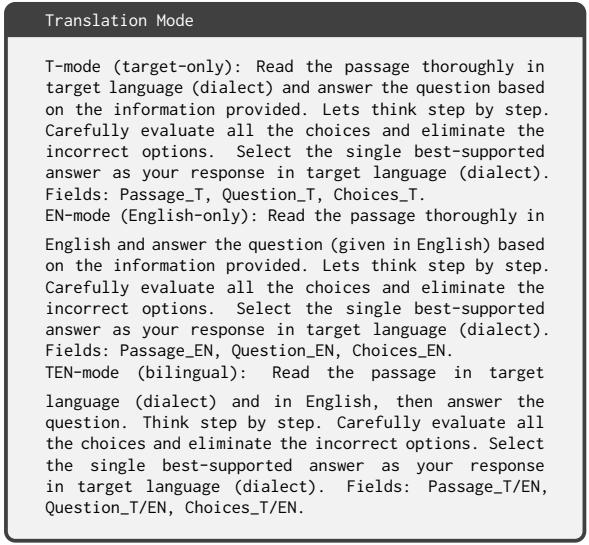

\scriptsize\ttfamily
\centering
\begin{tcolorbox}[title=Translation Mode]
T-mode (target-only):
Read the passage thoroughly in target language ({dialect}) and answer the question based on the information provided. Lets think step by step. Carefully evaluate all the choices and eliminate the incorrect options. Select the single best-supported answer as your response in target language ({dialect}).
Fields: Passage\_T, Question\_T, Choices\_T.\\
\smallskip
EN-mode (English-only):
Read the passage thoroughly in English and answer the question (given in English) based on the information provided. Lets think step by step. Carefully evaluate all the choices and eliminate the incorrect options. Select the single best-supported answer as your response in target language ({dialect}).
Fields: Passage\_EN, Question\_EN, Choices\_EN.\\
\smallskip
TEN-mode (bilingual):
Read the passage in target language ({dialect}) and in English, then answer the question. Think step by step. Carefully evaluate all the choices and eliminate the incorrect options. Select the single best-supported answer as your response in target language ({dialect}).
Fields: Passage\_T/EN, Question\_T/EN, Choices\_T/EN.
\end{tcolorbox}
\caption{Evaluation prompt templates (abridged).}
\label{fig:prompt_templates_appendix}
\end{figure}

\subsection{SeqGate Prompt Template}
\label{sec:seq_prompt}

Figure~\ref{fig:seq_prompt_example} shows the sequential bilingual reasoning template used for MMLU-ProX-Lite.
\begin{figure*}[!t]
\centering
\begin{tcolorbox}[title=Sequential bilingual reasoning prompt]
\ttfamily\footnotesize
Prompt = """You will reason in three phases and then answer strictly with [[0]], [[1]], ..., [[9]].\\
\smallskip
Phase 1: Target-only reasoning:\\
Read the target question and choices, think step by step in \{dialect\} about the correct choice.\\
Do not output the final answer yet.\\
\smallskip
Q(T):\\
\smallskip
C(T):\\
\smallskip
Phase 2: English-only reasoning:\\
Read the English question and choices and think step by step in English about the correct choice.\\
Again, do not output the final answer yet.\\
\smallskip
Q(EN):\\
C(EN):\\
\smallskip
Phase 3 -- Reconcile and decide:\\
Combine the insights from Phase 1 and Phase 2, choose the single best-supported option.\\
Strictly follow this format: [[0]], [[1]], ..., [[9]]. Do not provide any feedback.\\
\smallskip
Your Answer:\\
"""
\end{tcolorbox}
\caption{Sequential bilingual reasoning prompt used for MMLU-ProX-Lite.}
\label{fig:seq_prompt_example}
\end{figure*}

\section{Experiment and Full per-resource and per-model and per -dataset breakdown}
\label{sec:appendix_resource_family}
\label{app:full-results}
We include experimental details for reproducibility and also report the complete accuracy, confidence, ECE, and entropy results underlying the resource-tier comparisons in Table~\ref{tab:m2_m3_res_compare}.
\subsection{Experimental Details}
 In this section, we describe additional experimental
 details and list the LLM prompts employed. All
 experiments are conducted with LLaMA 3.1-8B Instruct and Qwen3-4B on an HPC server equipped
 with Nvidia A40 GPUs. For the Belebele dataset,
 from the full set of 126 languages, we select 9 languages to cover a diverse range of resource levels.
 For MMLU, we use the STEM subset corresponding to the same language variants as those chosen
 for Belebele.}
\subsection{Findings of RAAI: Complete Metrics}
Tables~\ref{tab:Appendix_module1_res_full_belebele}--\ref{tab:Appendix_module1_res_full_qwen_mmlu} report the static T, EN, and TEN controls.
\begin{table*}[!t]
\centering
\small
\resizebox{\textwidth}{!}{%
\begin{tabular}{l|cccc|cccc|cccc}
\toprule
Resource & \multicolumn{4}{c|}{T} & \multicolumn{4}{c|}{EN} & \multicolumn{4}{c}{TEN} \\
 & Acc$\uparrow$ & Conf$\uparrow$ & ECE$\downarrow$  & Ent$\downarrow$  & Acc$\uparrow$ & Conf$\uparrow$ & ECE$\downarrow$  & Ent$\downarrow$ & Acc$\uparrow$ & Conf$\uparrow$ & ECE$\downarrow$  & Ent$\downarrow$ \\
\midrule
High & 81.26 & \textbf{76.58}& 9.15 & \textbf{0.64}& \textbf{81.79}& 74.75 & 10.05 & 0.70 & 81.78 & 76.34 & \textbf{8.75}& 0.66 \\
Mid & \textbf{68.88}& 60.01 & 14.05 & 1.00 & 65.15 & 63.11 & 12.42 & 0.93 & 61.79 & \textbf{64.67}& \textbf{9.81}& \textbf{0.90} \\
Low & \textbf{44.59}& 39.38 & \textbf{12.36}& 1.29 & 44.30 & \textbf{47.36}& 12.81 & \textbf{1.20}& 44.30 & 45.67 & 14.05 & 1.22 \\

\bottomrule
\end{tabular}
}%
\caption{Static-mode full metrics by resource tier (Belebele, LLaMA). Bold denotes the best value within each resource-tier row for each metric (higher Acc/Conf, lower ECE/Ent).}
\label{tab:Appendix_module1_res_full_belebele}
\end{table*}

\begin{table*}[!t]
\centering
\small
\resizebox{\textwidth}{!}{%
\begin{tabular}{l|cccc|cccc|cccc}
\toprule
Resource & \multicolumn{4}{c|}{T} & \multicolumn{4}{c|}{EN} & \multicolumn{4}{c}{TEN} \\
 & Acc$\uparrow$ & Conf$\uparrow$ & ECE$\downarrow$  & Ent$\downarrow$  & Acc$\uparrow$ & Conf$\uparrow$ & ECE$\downarrow$  & Ent$\downarrow$ & Acc$\uparrow$ & Conf$\uparrow$ & ECE$\downarrow$  & Ent$\downarrow$ \\
\midrule
High & 34.65 & 38.78 & \textbf{11.02}& 1.77 & 31.02 & \textbf{40.50}& 13.19 & \textbf{1.74}& \textbf{38.84}& 33.43 & 12.16 & 1.92 \\
Mid & 25.64 & 33.70 & 13.13 & \textbf{1.91}& \textbf{35.19}& \textbf{34.13}& \textbf{10.44}& \textbf{1.91}& 31.57 & 27.27 & 11.67 & 2.05 \\
Low & 23.47 & 29.72 & \textbf{10.02}& 1.99 & 29.35 & \textbf{32.83}& 13.10 & \textbf{1.95}& \textbf{33.70}& 28.44 & 13.46 & 2.05 \\

\bottomrule
\end{tabular}
}%
\caption{Static-mode full metrics by resource tier (MMLU-ProX-Lite, LLaMA). Bold denotes the best value within each resource-tier row for each metric (higher Acc/Conf, lower ECE/Ent).}
\label{tab:Appendix_module1_res_full_mmlu}
\end{table*}

\begin{table*}[!t]
\centering
\small
\resizebox{\textwidth}{!}{%
\begin{tabular}{l|cccc|cccc|cccc}
\toprule
Resource & \multicolumn{4}{c|}{T} & \multicolumn{4}{c|}{EN} & \multicolumn{4}{c}{TEN} \\
 & Acc$\uparrow$ & Conf$\uparrow$ & ECE$\downarrow$  & Ent$\downarrow$  & Acc$\uparrow$ & Conf$\uparrow$ & ECE$\downarrow$  & Ent$\downarrow$ & Acc$\uparrow$ & Conf$\uparrow$ & ECE$\downarrow$  & Ent$\downarrow$ \\
\midrule
High & 80.28 & 82.77 & 9.14 & 0.55 & 87.73 & \textbf{85.78}& \textbf{7.43}& 0.50 & \textbf{88.69}& 85.20 & 8.25 & \textbf{0.48} \\
Mid & 66.28 & 63.51 & 10.67 & 0.95 & \textbf{86.55}& \textbf{87.87}& \textbf{8.64}& 0.43 & 79.86 & 86.81 & 10.92 & \textbf{0.40} \\
Low & 45.95 & 45.67 & 9.42 & 1.22 & \textbf{85.44}& \textbf{86.72}& \textbf{6.48}& \textbf{0.46}& 81.65 & 78.81 & 9.31 & 0.61 \\

\bottomrule
\end{tabular}
}%
\caption{Static-mode full metrics by resource tier (Belebele, Qwen). Bold denotes the best value within each resource-tier row for each metric (higher Acc/Conf, lower ECE/Ent).}
\label{tab:Appendix_module1_res_full_qwen_belebele}
\end{table*}

\begin{table*}[!t]
\centering
\small
\resizebox{\textwidth}{!}{%
\begin{tabular}{l|cccc|cccc|cccc}
\toprule
Resource & \multicolumn{4}{c|}{T} & \multicolumn{4}{c|}{EN} & \multicolumn{4}{c}{TEN} \\
 & Acc$\uparrow$ & Conf$\uparrow$ & ECE$\downarrow$  & Ent$\downarrow$  & Acc$\uparrow$ & Conf$\uparrow$ & ECE$\downarrow$  & Ent$\downarrow$ & Acc$\uparrow$ & Conf$\uparrow$ & ECE$\downarrow$  & Ent$\downarrow$ \\
\midrule
High & 37.78 & 45.37 & 13.83 & 1.71 & 36.27 & 44.47 & \textbf{13.02}& 1.71 & \textbf{45.12}& \textbf{49.33}& 15.06 & \textbf{1.58} \\
Mid & 32.67 & 40.35 & \textbf{13.82}& 1.79 & 34.52 & 43.22 & 16.69 & 1.73 & \textbf{46.38}& \textbf{55.70}& 21.51 & \textbf{1.41} \\
Low & 28.57 & 38.15 & \textbf{12.81}& 1.83 & 35.87 & 44.28 & 18.24 & 1.72 & \textbf{44.57}& \textbf{55.30}& 17.63 & \textbf{1.43} \\

\bottomrule
\end{tabular}
}%
\caption{Static-mode full metrics by resource tier (MMLU-ProX-Lite, Qwen). Bold denotes the best value within each resource-tier row for each metric (higher Acc/Conf, lower ECE/Ent).}
\label{tab:Appendix_module1_res_full_qwen_mmlu}
\end{table*}

\subsection{\textsc{RouteGate}: Full Metrics}
\label{sec:RouteGate}
We report accuracy, confidence, ECE, and entropy by resource tier for T, TEN, and the adaptive method.

\begin{table*}[!t]
\centering
\small
\resizebox{\textwidth}{!}{%
\begin{tabular}{l|cccc|cccc|cccc}
\toprule
Resource & \multicolumn{4}{c|}{T} & \multicolumn{4}{c|}{TEN} & \multicolumn{4}{c}{\textsc{RouteGate}} \\
 & Acc$\uparrow$ & Conf$\uparrow$ & ECE$\downarrow$  & Ent$\downarrow$  & Acc$\uparrow$ & Conf$\uparrow$ & ECE$\downarrow$  & Ent$\downarrow$ & Acc$\uparrow$ & Conf$\uparrow$ & ECE$\downarrow$  & Ent$\downarrow$ \\
\midrule
High & 81.26 & 76.58 & 9.15 & 0.64 & 83.23 & 78.20 & 8.55 & 0.61 & \textbf{83.58} & \textbf{79.92} & \textbf{5.44} & \textbf{0.59} \\
Mid & 68.88 & 60.01 & 14.05 & 1.00 & 71.52 & 68.11 & 11.10 & 0.82 & \textbf{73.19} & \textbf{74.00} & \textbf{6.56} & \textbf{0.74} \\
Low & 44.59 & 39.38 & 12.36 & 1.29 & \textbf{55.22} & 52.58 & 10.83 & 1.10 & 55.19 & \textbf{58.90} & \textbf{7.71} & \textbf{1.01} \\
\bottomrule
\end{tabular}
}%

\caption{\textsc{RouteGate} full metrics by resource tier (Belebele, LLaMA). Bold denotes the best value within each resource-tier row for each metric (higher Acc/Conf, lower ECE/Ent).}
\label{tab:Appendix_module2_res_full_llama_belebele}
\end{table*}

\begin{table*}[!t]
\centering
\small
\resizebox{\textwidth}{!}{%
\begin{tabular}{l|cccc|cccc|cccc}
\toprule
Resource & \multicolumn{4}{c|}{T} & \multicolumn{4}{c|}{TEN} & \multicolumn{4}{c}{\textsc{RouteGate}} \\
 & Acc$\uparrow$ & Conf$\uparrow$ & ECE$\downarrow$  & Ent$\downarrow$  & Acc$\uparrow$ & Conf$\uparrow$ & ECE$\downarrow$  & Ent$\downarrow$ & Acc$\uparrow$ & Conf$\uparrow$ & ECE$\downarrow$  & Ent$\downarrow$ \\
\midrule
High & 34.65 & 38.78 & 11.02 & \textbf{1.77} & 31.29 & \textbf{39.07} & 13.92 & 1.78 & \textbf{35.92} & 34.28 & 5.42 & 1.91 \\
Mid & 25.64 & {33.70} & 13.13 & \textbf{1.91} & \textbf{30.75} & 31.89 & 10.09 & 1.96 & 27.95 & \textbf{36.62} & 3.38 & 2.07 \\
Low & 23.47 & 29.72 & 10.02 & \textbf{1.99} & \textbf{31.10} & {30.39} & 13.04 & 2.00 & 28.62 & \textbf{37.57} & 4.00 & 2.06 \\
\bottomrule
\end{tabular}
}%

\caption{\textsc{RouteGate} full metrics by resource tier (MMLU-ProX-Lite, LLaMA). Bold denotes the best value within each resource-tier row for each metric (higher Acc/Conf, lower ECE/Ent).}
\label{tab:Appendix_module2_res_full_llama_mmlu}
\end{table*}

\begin{table*}[!t]
\centering
\small
\resizebox{\textwidth}{!}{%
\begin{tabular}{l|cccc|cccc|cccc}
\toprule
Resource & \multicolumn{4}{c|}{T} & \multicolumn{4}{c|}{TEN} & \multicolumn{4}{c}{\textsc{RouteGate}} \\
 & Acc$\uparrow$ & Conf$\uparrow$ & ECE$\downarrow$  & Ent$\downarrow$  & Acc$\uparrow$ & Conf$\uparrow$ & ECE$\downarrow$  & Ent$\downarrow$ & Acc$\uparrow$ & Conf$\uparrow$ & ECE$\downarrow$  & Ent$\downarrow$ \\
\midrule
High & 80.28 & 82.77 & 9.14 & 0.55 & \textbf{84.21} & 83.88 & 8.87 & \textbf{0.52} & 83.77 & \textbf{84.70} & \textbf{5.19} & \textbf{0.52} \\
Mid & 66.28 & 63.51 & 10.67 & 0.95 & \textbf{72.84} & 73.38 & \textbf{9.98} & 0.73 & 69.33 & \textbf{79.72} & 10.76 & \textbf{0.61} \\
Low & 45.95 & 45.67 & \textbf{9.42} & 1.22 & {56.44} & 59.08 & 10.92 & 0.99 & \textbf{71.14} & \textbf{62.78} & 13.19 & \textbf{0.96} \\
\bottomrule
\end{tabular}
}%

\caption{\textsc{RouteGate} full metrics by resource tier (Belebele, Qwen). Bold denotes the best value within each resource-tier row for each metric (higher Acc/Conf, lower ECE/Ent).}
\label{tab:Appendix_module2_res_full_qwen_belebele}
\end{table*}

\begin{table*}[!t]
\centering
\small
\resizebox{\textwidth}{!}{%
\begin{tabular}{l|cccc|cccc|cccc}
\toprule
Resource & \multicolumn{4}{c|}{T} & \multicolumn{4}{c|}{TEN} & \multicolumn{4}{c}{\textsc{RouteGate}} \\
 & Acc$\uparrow$ & Conf$\uparrow$ & ECE$\downarrow$  & Ent$\downarrow$  & Acc$\uparrow$ & Conf$\uparrow$ & ECE$\downarrow$  & Ent$\downarrow$ & Acc$\uparrow$ & Conf$\uparrow$ & ECE$\downarrow$  & Ent$\downarrow$ \\
\midrule
High & 37.78 & 45.37 & 13.83 & 1.71 & 36.83 & 42.84 & 13.28 & 1.75 & \textbf{42.04} & \textbf{46.45} & 8.44 & \textbf{1.65} \\
Mid & 32.67 & 40.35 & 13.82 & 1.79 & 34.66 & 41.77 & 14.37 & 1.77 & \textbf{41.78} & \textbf{47.28} & 11.09 & \textbf{1.61} \\
Low & 28.57 & 38.15 & 12.81 & 1.83 & 34.86 & 41.48 & 16.39 & 1.78 & \textbf{38.04} & \textbf{47.87} & 10.65 & \textbf{1.61} \\
\bottomrule
\end{tabular}
}%

\caption{\textsc{RouteGate} full metrics by resource tier (MMLU-ProX-Lite, Qwen). Bold denotes the best value within each resource-tier row for each metric (higher Acc/Conf, lower ECE/Ent).}
\label{tab:Appendix_module2_res_full_qwen_mmlu}
\end{table*}

\subsection{\textsc{SeqGate}: Full Metrics}
\label{sec:SEQmetric}

We report accuracy, confidence, ECE, and entropy by resource tier for T, TEN, and the adaptive method.

\begin{table*}[!t]
\centering
\small
\resizebox{\textwidth}{!}{%
\begin{tabular}{l|cccc|cccc|cccc}
\toprule
Resource & \multicolumn{4}{c|}{T} & \multicolumn{4}{c|}{TEN} & \multicolumn{4}{c}{\textsc{SeqGate}}  \\
 & Acc$\uparrow$ & Conf$\uparrow$ & ECE$\downarrow$  & Ent$\downarrow$  & Acc$\uparrow$ & Conf$\uparrow$ & ECE$\downarrow$  & Ent$\downarrow$ & Acc$\uparrow$ & Conf$\uparrow$ & ECE$\downarrow$  & Ent$\downarrow$ \\
\midrule
High & 81.26 & 76.58 & 9.15 & 0.64 & \textbf{84.79} & 78.05 & 8.43 & 0.62 & 82.58 & \textbf{78.52} & \textbf{6.52} & \textbf{0.61} \\
Mid & 68.88 & 60.01 & 14.05 & 1.00 & \textbf{72.87} & 64.70 & 11.99 & 0.91 & 67.68 & \textbf{66.93} & \textbf{10.62} & \textbf{0.86} \\
Low & 44.59 & 39.38 & 12.36 & 1.29 & 50.68 & 47.61 & \textbf{6.47} & 1.18 & \textbf{69.70} & \textbf{66.13} & 11.34 & \textbf{0.88} \\
\bottomrule
\end{tabular}
}%

\caption{\textsc{SeqGate} full metrics by resource tier (Belebele, LLaMA). Bold denotes the best value within each resource-tier row for each metric (higher Acc/Conf, lower ECE/Ent).}
\label{tab:Appendix_module3_res_full_llama_belebele}
\end{table*}

\begin{table*}[!t]
\centering
\small
\resizebox{\textwidth}{!}{%
\begin{tabular}{l|cccc|cccc|cccc}
\toprule
Resource & \multicolumn{4}{c|}{T} & \multicolumn{4}{c|}{TEN} & \multicolumn{4}{c}{\textsc{SeqGate}} \\
 & Acc$\uparrow$ & Conf$\uparrow$ & ECE$\downarrow$  & Ent$\downarrow$  & Acc$\uparrow$ & Conf$\uparrow$ & ECE$\downarrow$  & Ent$\downarrow$ & Acc$\uparrow$ & Conf$\uparrow$ & ECE$\downarrow$  & Ent$\downarrow$ \\
\midrule
High & \textbf{34.65} & \textbf{38.78} & 11.02 & \textbf{1.77} & 32.16 & 35.29 & 12.51 & 1.88 & 33.76 & 34.83 & \textbf{10.59} & 1.88 \\
Mid & 25.64 & \textbf{33.70} & 13.13 & \textbf{1.91} & 24.67 & 27.17 & 13.17 & 2.06 & \textbf{33.73} & 24.61 & \textbf{8.76} & 2.12 \\
Low & 23.47 & \textbf{29.72} & 10.02 & \textbf{1.99} & \textbf{34.12} & 28.46 & \textbf{8.83} & 2.04 & \textbf{34.12} & 25.32 & 9.16 & 2.10 \\
\bottomrule
\end{tabular}
}%

\caption{\textsc{SeqGate} full metrics by resource tier (MMLU-ProX-Lite, LLaMA). Bold denotes the best value within each resource-tier row for each metric (higher Acc/Conf, lower ECE/Ent).}
\label{tab:Appendix_module3_res_full_llama_mmlu}
\end{table*}

\begin{table*}[!t]
\centering
\small
\resizebox{\textwidth}{!}{%
\begin{tabular}{l|cccc|cccc|cccc}
\toprule
Resource & \multicolumn{4}{c|}{T} & \multicolumn{4}{c|}{TEN} & \multicolumn{4}{c}{\textsc{SeqGate}}  \\
 & Acc$\uparrow$ & Conf$\uparrow$ & ECE$\downarrow$  & Ent$\downarrow$  & Acc$\uparrow$ & Conf$\uparrow$ & ECE$\downarrow$  & Ent$\downarrow$ & Acc$\uparrow$ & Conf$\uparrow$ & ECE$\downarrow$  & Ent$\downarrow$ \\
\midrule
High & 80.28 & 82.77 & 9.14 & 0.55 & 81.47 & \textbf{83.41} & 8.17 & \textbf{0.52} & \textbf{88.54} & 83.36 & \textbf{7.69} & 0.54 \\
Mid & 66.28 & 63.51 & 10.67 & 0.95 & \textbf{69.14} & 68.48 & \textbf{8.74} & 0.84 & 62.96 & \textbf{71.76} & 11.68 & \textbf{0.77} \\
Low & 45.95 & 45.67 & 9.42 & 1.22 & 49.32 & 51.00 & 10.70 & 1.13 & \textbf{83.64} & \textbf{71.71} & \textbf{8.35} & \textbf{0.78} \\
\bottomrule
\end{tabular}
}%

\caption{\textsc{SeqGate} full metrics by resource tier (Belebele, Qwen). Bold denotes the best value within each resource-tier row for each metric (higher Acc/Conf, lower ECE/Ent).}
\label{tab:Appendix_module3_res_full_qwen_belebele}
\end{table*}

\begin{table*}[!t]
\centering
\small
\resizebox{\textwidth}{!}{%
\begin{tabular}{l|cccc|cccc|cccc}
\toprule
Resource & \multicolumn{4}{c|}{T} & \multicolumn{4}{c|}{TEN} & \multicolumn{4}{c}{\textsc{SeqGate}} \\
 & Acc$\uparrow$ & Conf$\uparrow$ & ECE$\downarrow$  & Ent$\downarrow$  & Acc$\uparrow$ & Conf$\uparrow$ & ECE$\downarrow$  & Ent$\downarrow$ & Acc$\uparrow$ & Conf$\uparrow$ & ECE$\downarrow$  & Ent$\downarrow$ \\
\midrule
High & 37.78 & \textbf{45.37} & 13.83 & \textbf{1.71} & {41.00} & 44.76 & 17.77 & \textbf{1.71} & \textbf{47.13} & 42.47 & \textbf{10.41} & 1.77 \\
Mid & 32.67 & 40.35 & 13.82 & 1.79 & {38.70} & \textbf{44.01} & \textbf{13.52} & \textbf{1.71} & \textbf{51.64} & 37.57 & 14.04 & 1.87 \\
Low & 28.57 & 38.15 & \textbf{12.81} & 1.83 & {36.96} & \textbf{45.04} & 17.72 & \textbf{1.69} & \textbf{51.36} & 37.89 & 13.49 & 1.86 \\
\bottomrule
\end{tabular}
}%

\caption{\textsc{SeqGate} full metrics by resource tier (MMLU-ProX-Lite, Qwen). Bold denotes the best value within each resource-tier row for each metric (higher Acc/Conf, lower ECE/Ent).}
\label{tab:Appendix_module3_res_full_qwen_mmlu}
\end{table*}

\section{Expanded Diagnostic Analyses}
\label{app:diagnostics}
\subsection{Full Layer-Wise Curves}
\label{sec:appendix_layer_curves}
The full-resolution plots complement Figure~\ref{fig:layer_all_vs_choices} by showing per-language accuracy, confidence, ECE, and entropy trajectories for representative question- and choice-translation settings.
\paragraph{Question translation.}
Figures~\ref{fig:layer_acc_appendix_QB}--\ref{fig:layer_entropy_appendix_QB} compare Qwen on Belebele and LLaMA on MMLU-ProX-Lite under question translation.
\begin{figure*}[!t]
  \centering
  \begin{subfigure}[!t]{0.48\textwidth}
    \centering
    \includegraphics[width=\linewidth]{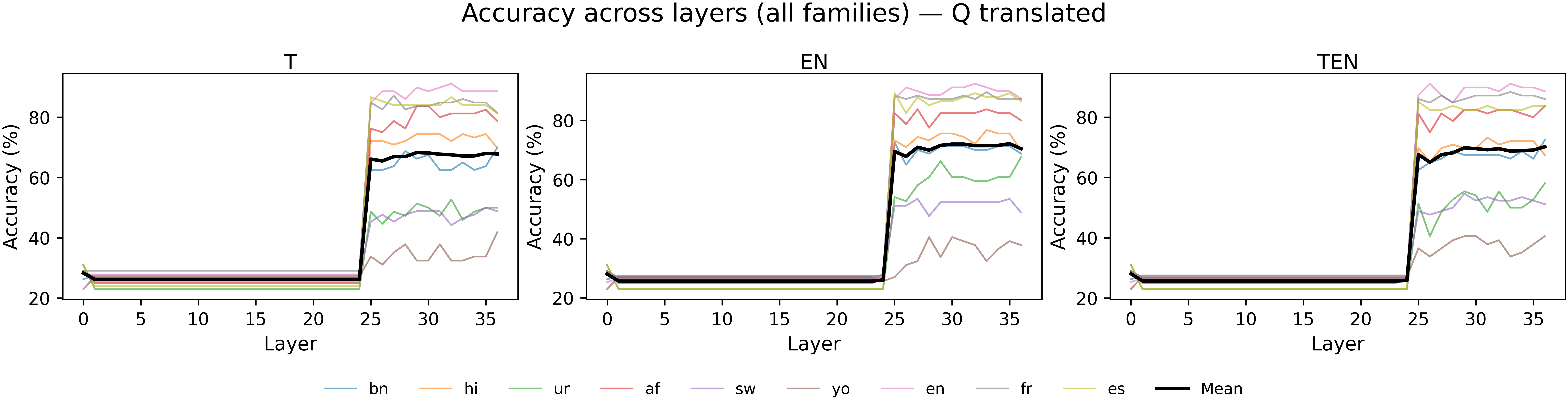}
    \caption{Qwen on Belebele.}
    \label{fig:QBQA}
  \end{subfigure}
  \hfill
  \begin{subfigure}[!t]{0.48\textwidth}
    \centering
    \includegraphics[width=\linewidth]{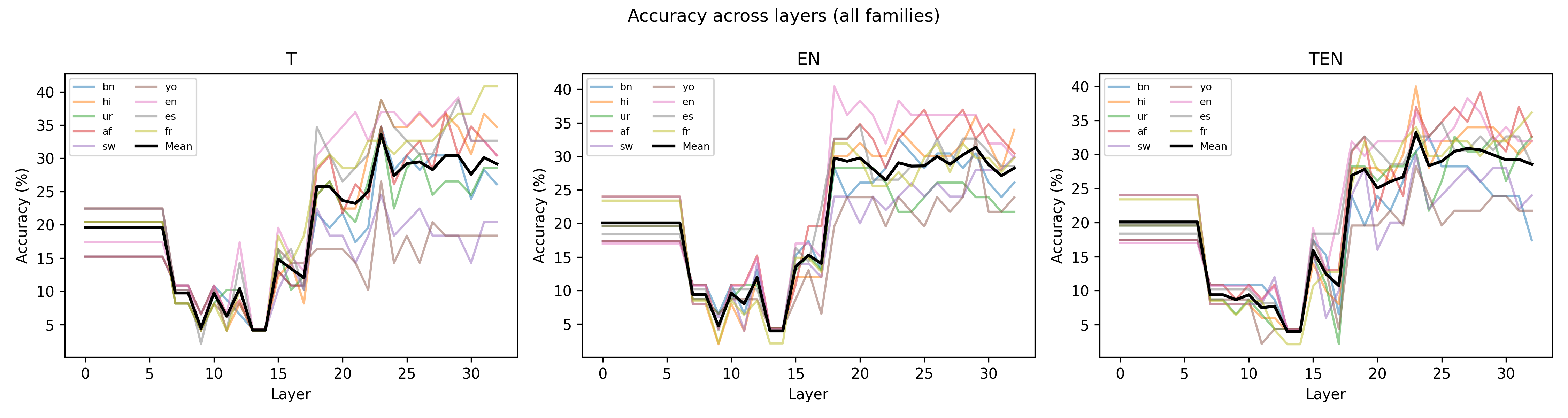}
    \caption{LLaMA on MMLU-ProX-Lite.}
    \label{fig:LMQA}
  \end{subfigure}
  \caption{Layer-wise accuracy under question translation.}
  \label{fig:layer_acc_appendix_QB}
\end{figure*}

\begin{figure*}[!t]
  \centering
  \begin{subfigure}[!t]{0.48\textwidth}
    \centering
    \includegraphics[width=\linewidth]{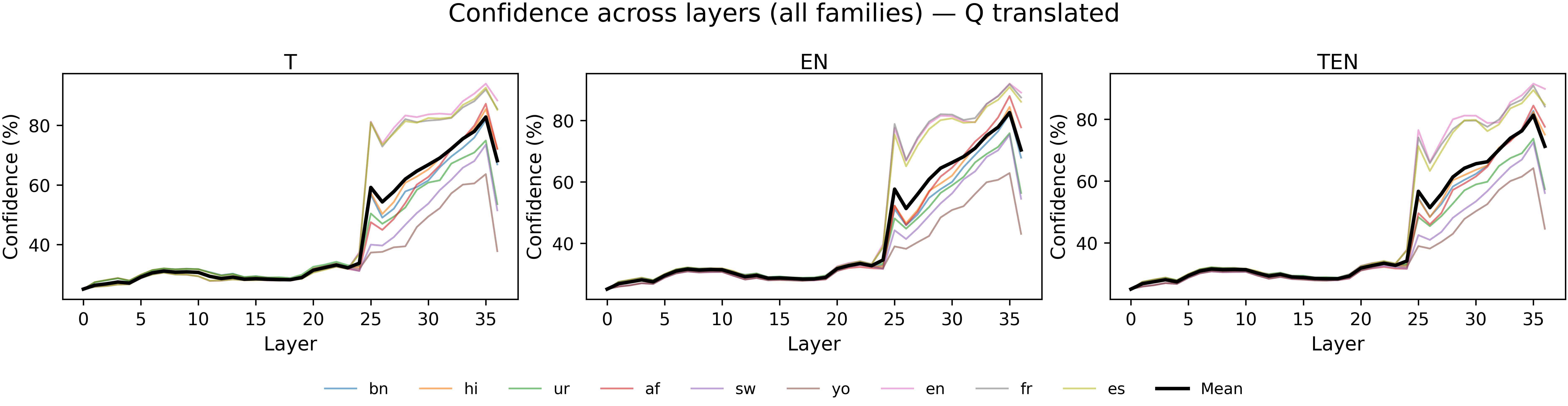}
    \caption{Qwen on Belebele.}
    \label{fig:QBQC}
  \end{subfigure}
  \hfill
  \begin{subfigure}[!t]{0.48\textwidth}
    \centering
    \includegraphics[width=\linewidth]{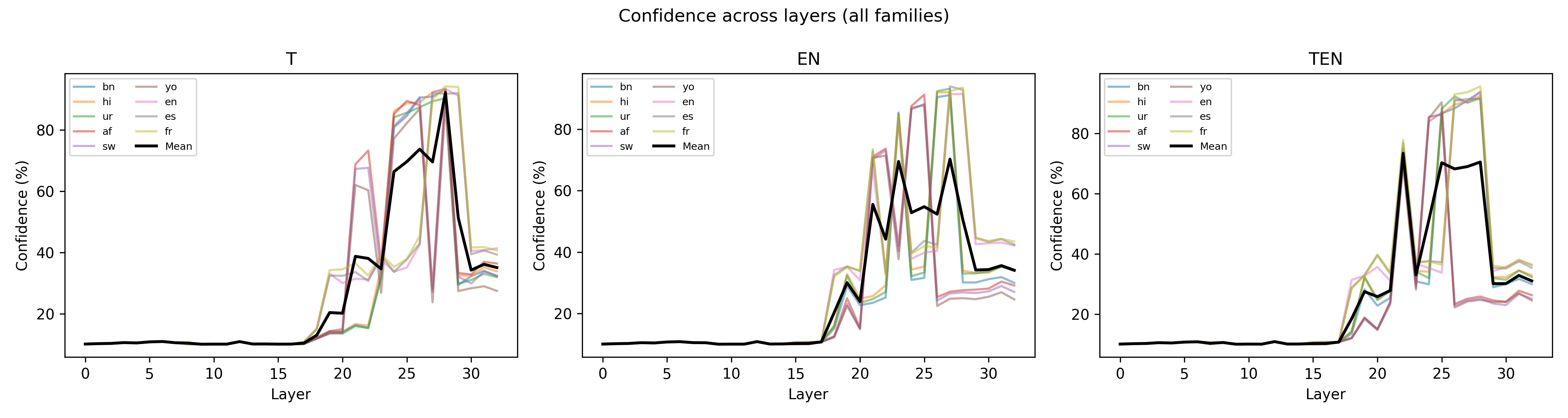}
    \caption{LLaMA on MMLU-ProX-Lite.}
    \label{fig:LMQC}
  \end{subfigure}
  \caption{Layer-wise confidence under question translation.}
  \label{fig:layer_conf_appendix_QB}
\end{figure*}
\begin{figure*}[!t]
  \centering
  \begin{subfigure}[!t]{0.48\textwidth}
    \centering
    \includegraphics[width=\linewidth]{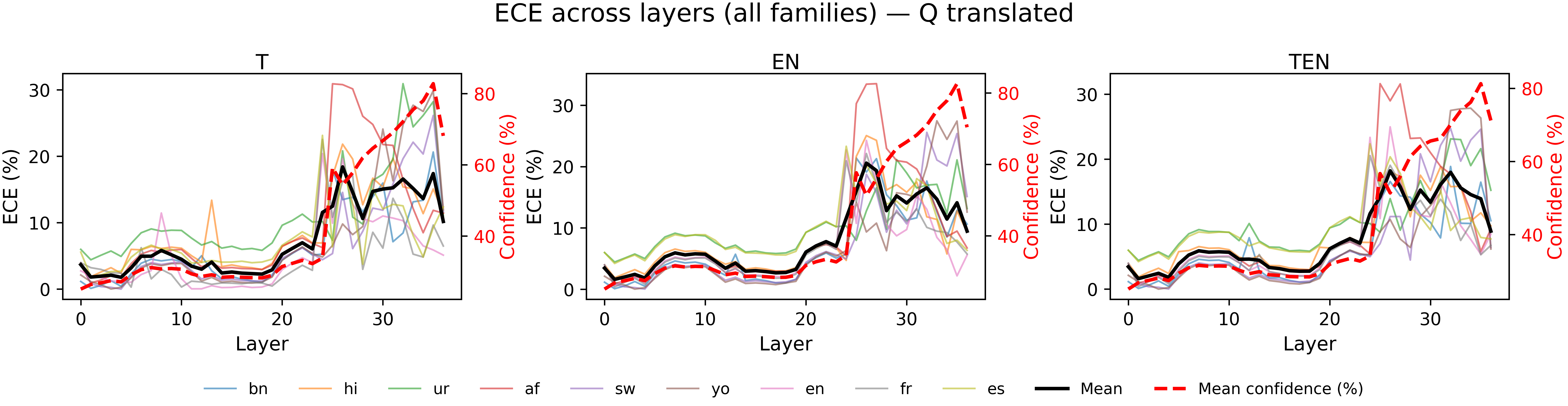}
    \caption{Qwen on Belebele.}
    \label{fig:QBQEce}
  \end{subfigure}
  \hfill
  \begin{subfigure}[!t]{0.48\textwidth}
    \centering
    \includegraphics[width=\linewidth]{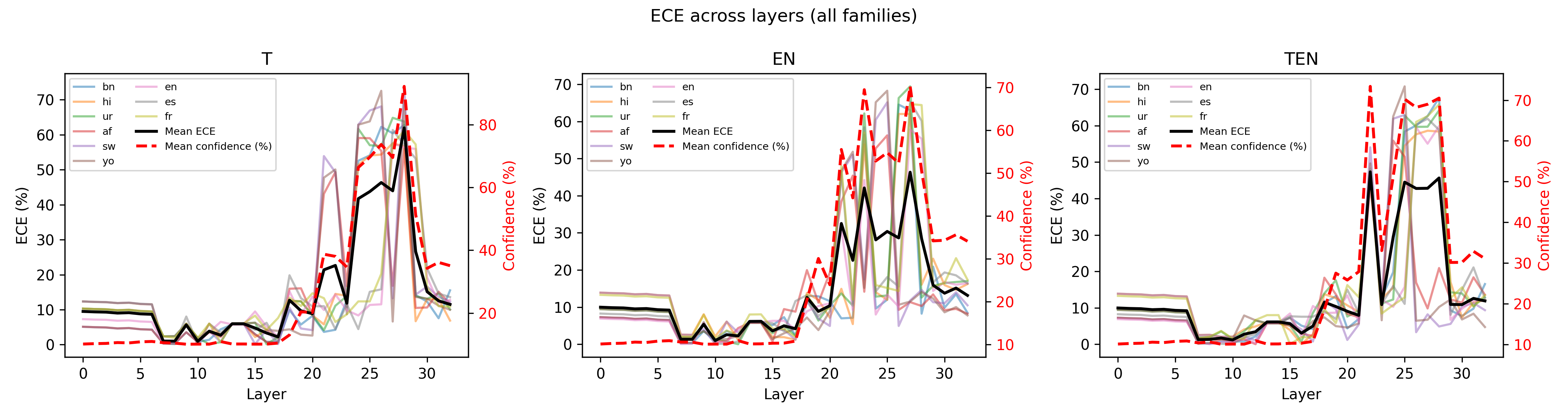}
    \caption{LLaMA on MMLU-ProX-Lite.}
    \label{fig:LMQEce}
  \end{subfigure}
  \caption{Layer-wise ECE under question translation.}
  \label{fig:layer_ece_appendix_QB}
\end{figure*}

\begin{figure*}[!t]
  \centering
  \begin{subfigure}[!t]{0.48\textwidth}
    \centering
    \includegraphics[width=\linewidth]{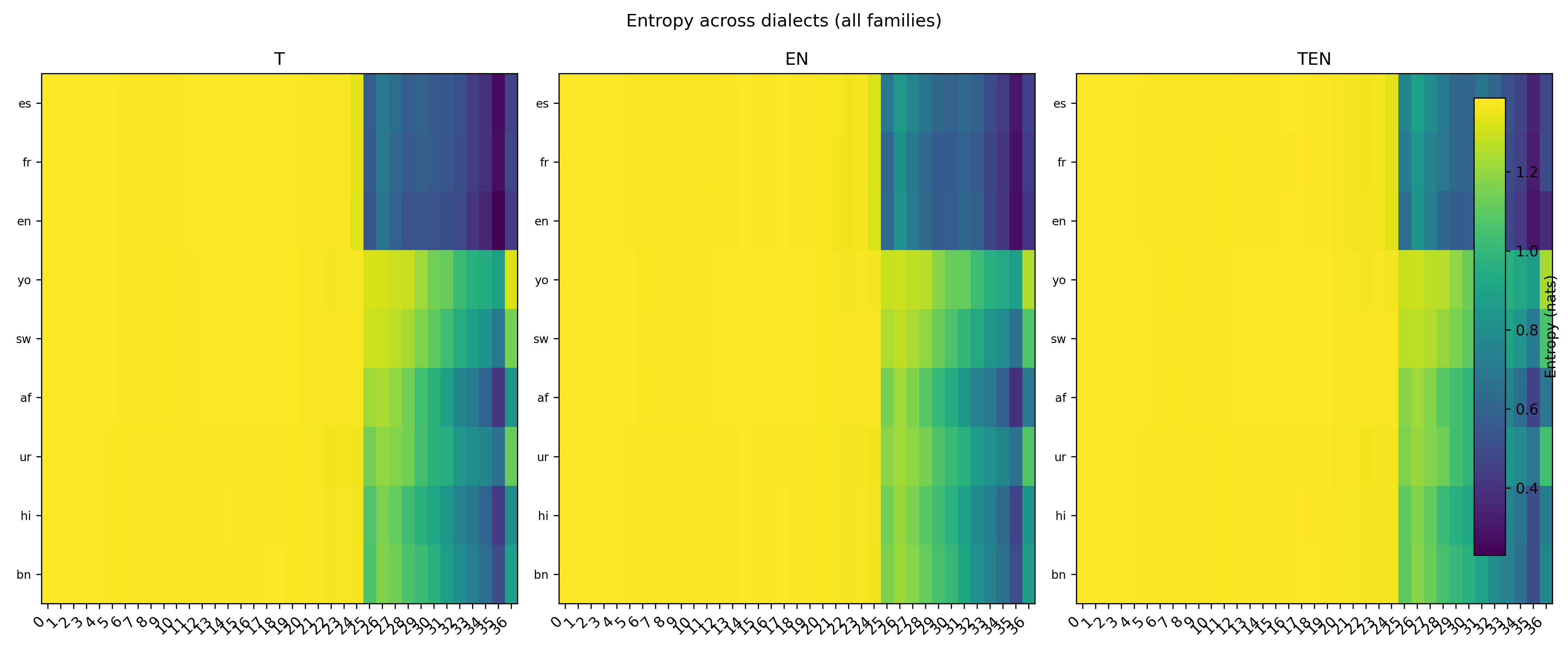}
    \caption{Qwen on Belebele.}
    \label{fig:QBQE}
  \end{subfigure}
  \hfill
  \begin{subfigure}[!t]{0.48\textwidth}
    \centering
    \includegraphics[width=\linewidth]{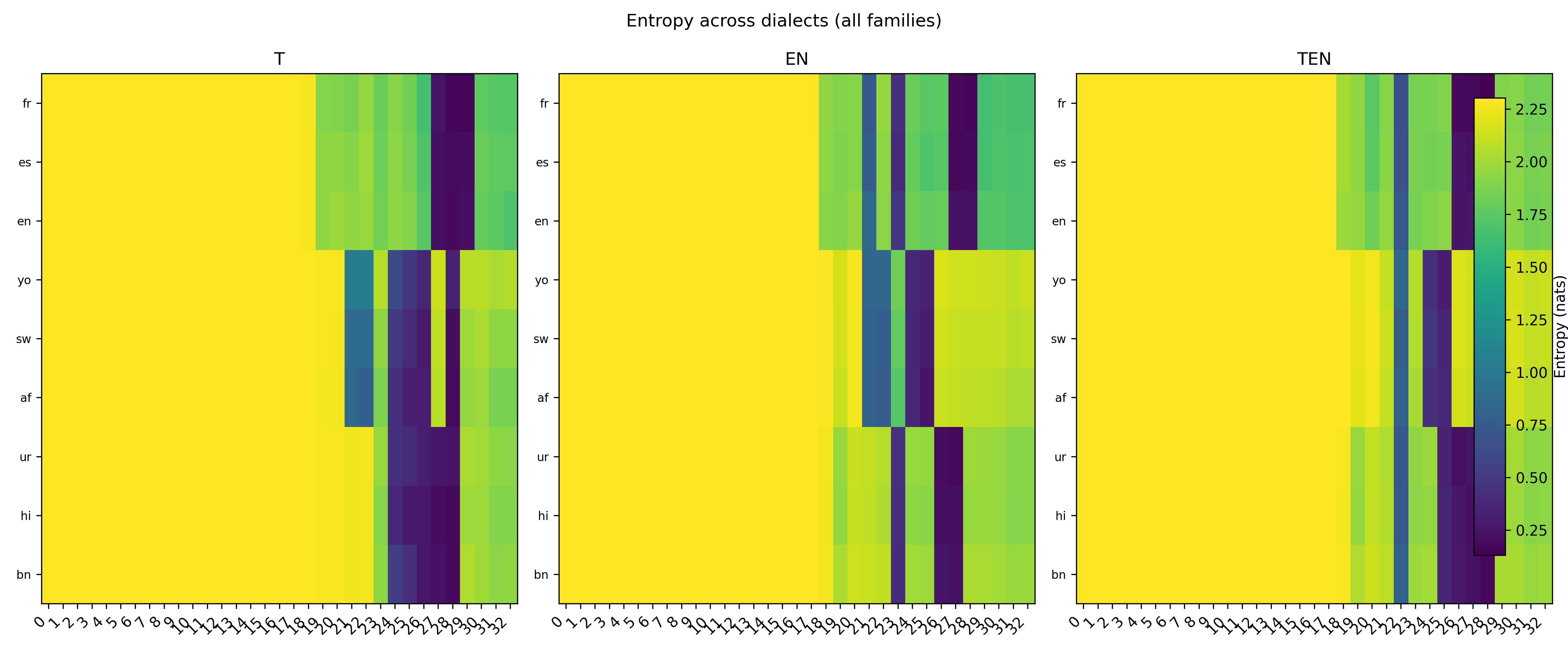}
    \caption{LLaMA on MMLU-ProX-Lite.}
    \label{fig:LMQE}
  \end{subfigure}
  \caption{Layer-wise entropy under question translation.}
  \label{fig:layer_entropy_appendix_QB}
\end{figure*}

\paragraph{Choice translation.}
Figures~\ref{fig:layer_acc_appendix_QM}--\ref{fig:layer_entropy_appendix_QM} show the corresponding choice-translation comparisons.
\begin{figure*}[!t]
  \centering
  \begin{subfigure}[!t]{0.48\textwidth}
    \centering
    \includegraphics[width=\linewidth]{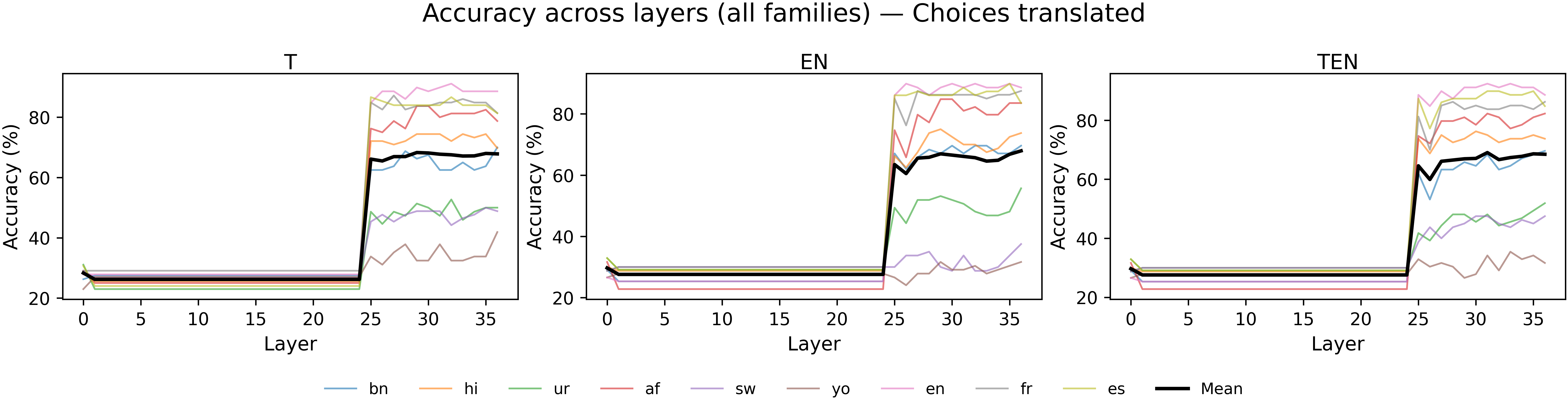}
    \caption{Qwen on Belebele.}
    \label{fig:QMAC}
  \end{subfigure}
  \hfill
  \begin{subfigure}[!t]{0.48\textwidth}
    \centering
    \includegraphics[width=\linewidth]{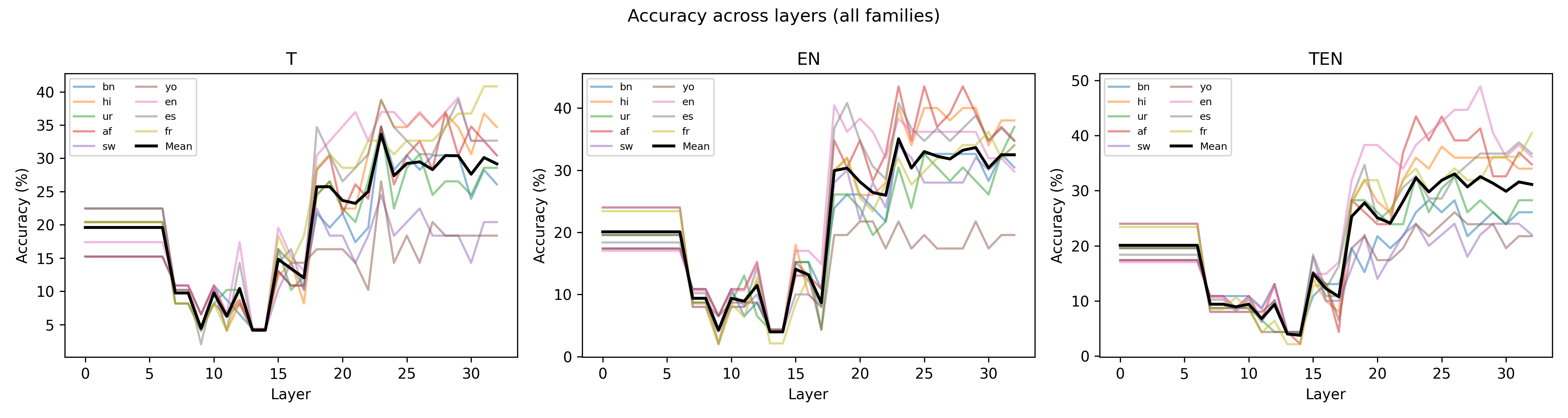}
    \caption{LLaMA on MMLU-ProX-Lite.}
    \label{fig:LMAC}
  \end{subfigure}
  \caption{Layer-wise accuracy under choice translation.}
  \label{fig:layer_acc_appendix_QM}
\end{figure*}
\begin{figure*}[!t]
  \centering
  \begin{subfigure}[!t]{0.48\textwidth}
    \centering
    \includegraphics[width=\linewidth]{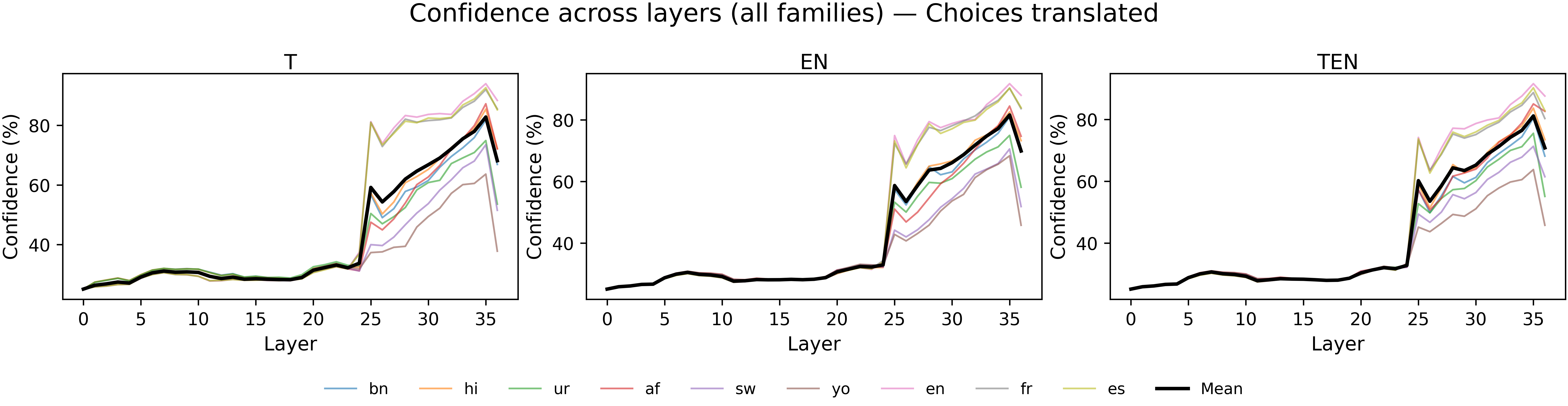}
    \caption{Qwen on Belebele.}
    \label{fig:QBC}
  \end{subfigure}
  \hfill
  \begin{subfigure}[!t]{0.48\textwidth}
    \centering
    \includegraphics[width=\linewidth]{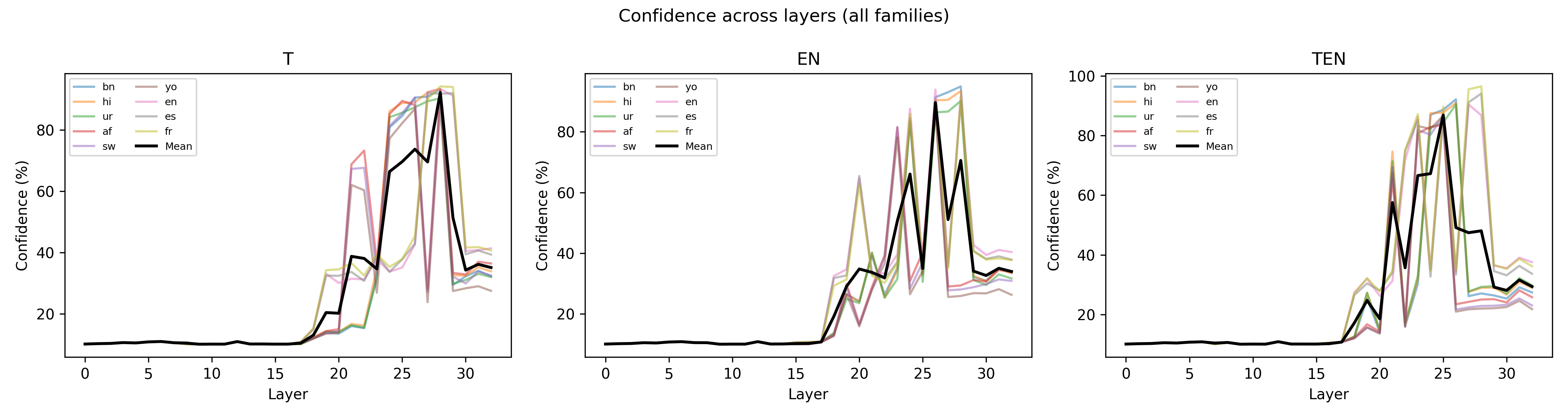}
    \caption{LLaMA on MMLU-ProX-Lite.}
    \label{fig:layer-c}
  \end{subfigure}
  \caption{Layer-wise confidence under choice translation.}
  \label{fig:layer_conf_appendix_QM}
\end{figure*}
\begin{figure*}[!t]
  \centering
  \begin{subfigure}[!t]{0.48\textwidth}
    \centering
    \includegraphics[width=\linewidth]{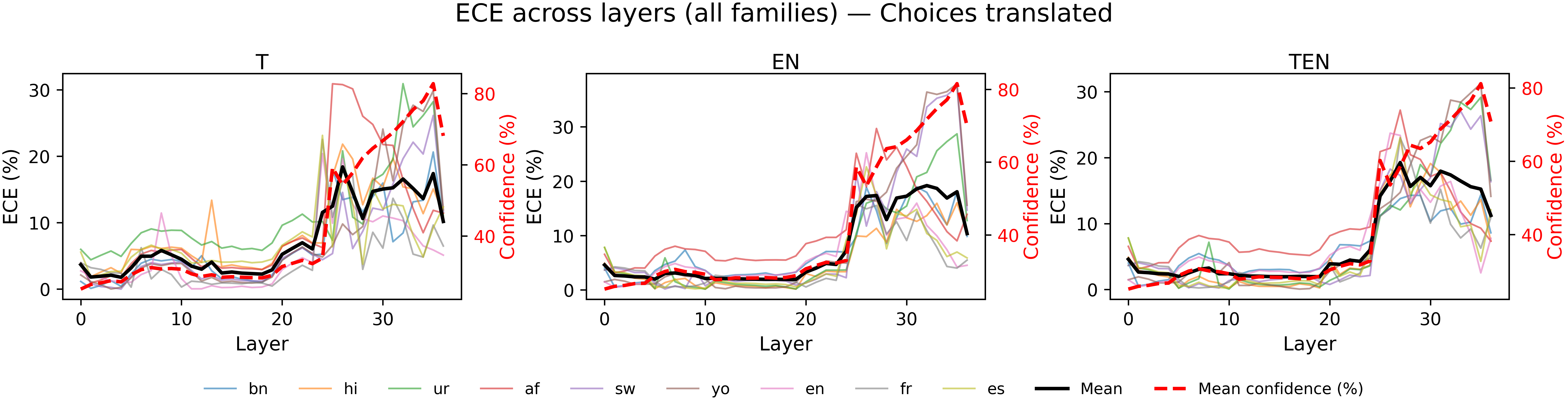}
    \caption{Qwen on Belebele.}
  \end{subfigure}
  \hfill
  \begin{subfigure}[!t]{0.48\textwidth}
    \centering
    \includegraphics[width=\linewidth]{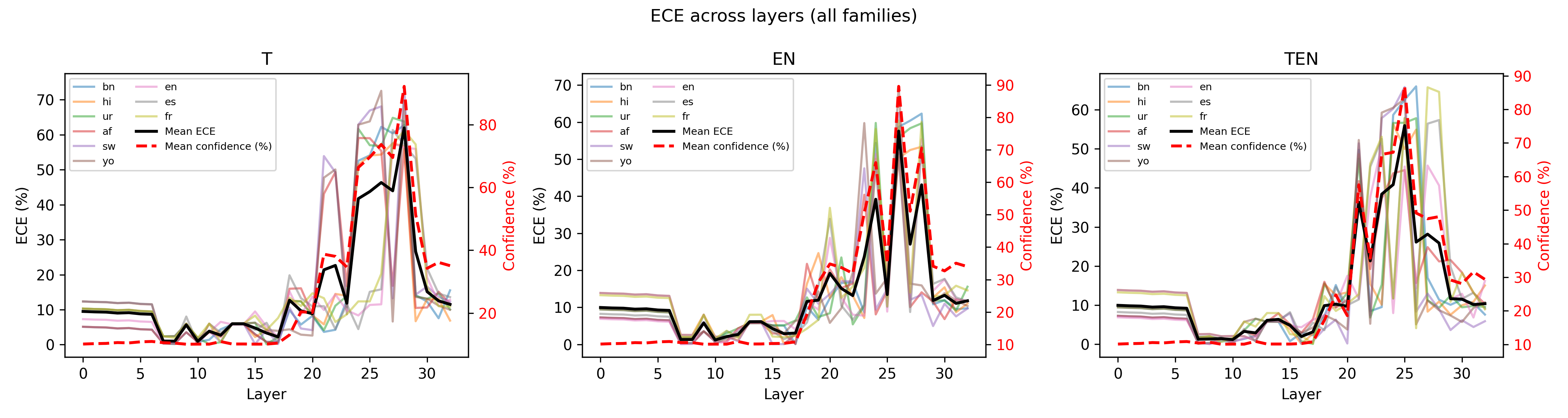}
    \caption{LLaMA on MMLU-ProX-Lite.}
    \label{fig:raai-m3}
  \end{subfigure}
  \caption{Layer-wise ECE under choice translation.}
  \label{fig:layer_ece_appendix_QM}
\end{figure*}

\begin{figure*}[!t]
  \centering
  \begin{subfigure}[!t]{0.48\textwidth}
    \centering
    \includegraphics[width=\linewidth]{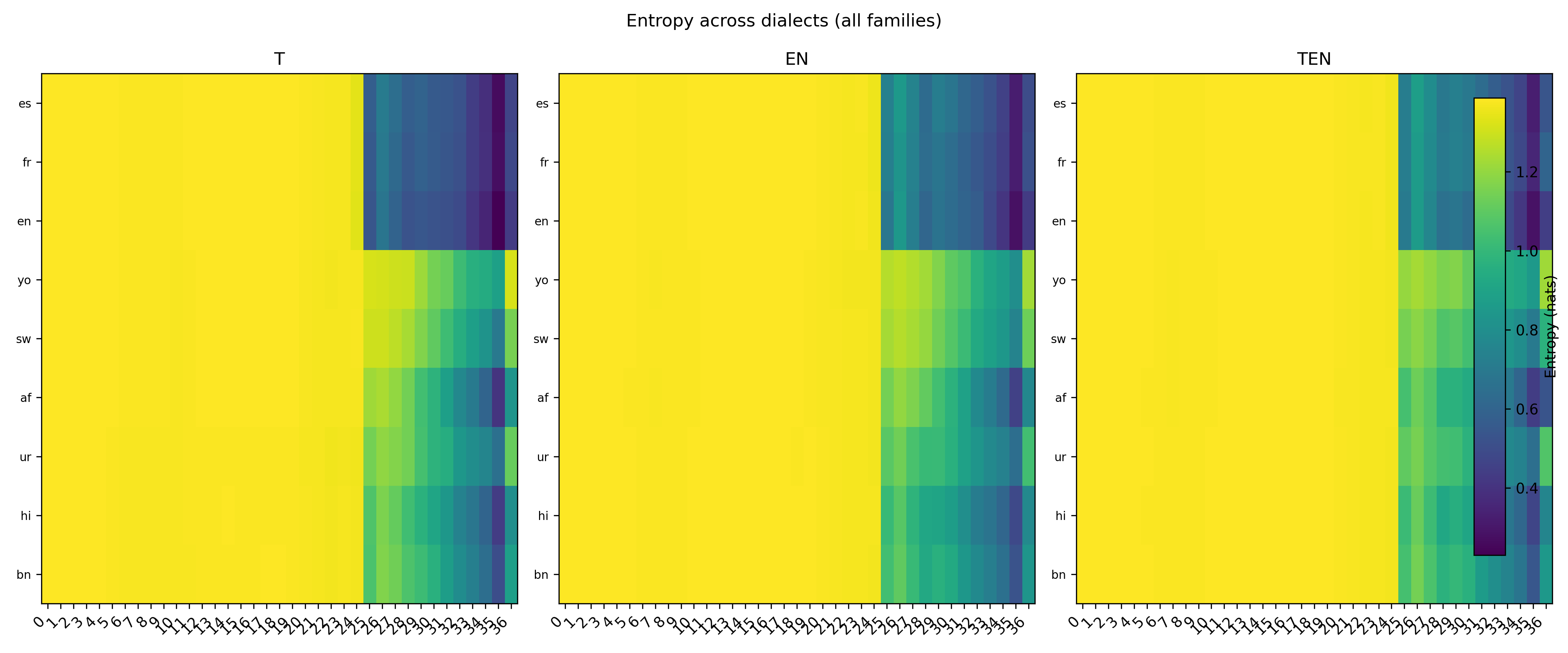}
    \caption{Qwen on Belebele.}
  \end{subfigure}
  \hfill
  \begin{subfigure}[!t]{0.48\textwidth}
    \centering
    \includegraphics[width=\linewidth]{latex/results/llama_mmlu/module1/Q/entropy_heatmap_all_families.png}
    \caption{LLaMA on MMLU-ProX-Lite.}
  \end{subfigure}
  \caption{Layer-wise entropy under choice translation.}
  \label{fig:layer_entropy_appendix_QM}
\end{figure*}

\subsection{Root-Cause Analysis}
\label{sec:root-cause}
Figures~\ref{fig:QLRC}--\ref{fig:QBRC} provide the remaining correctness-taxonomy breakdowns for LLaMA and Qwen on MMLU-ProX-Lite and for Qwen on Belebele.
\begin{figure*}[!t]
  \centering
\includegraphics[width=\linewidth]{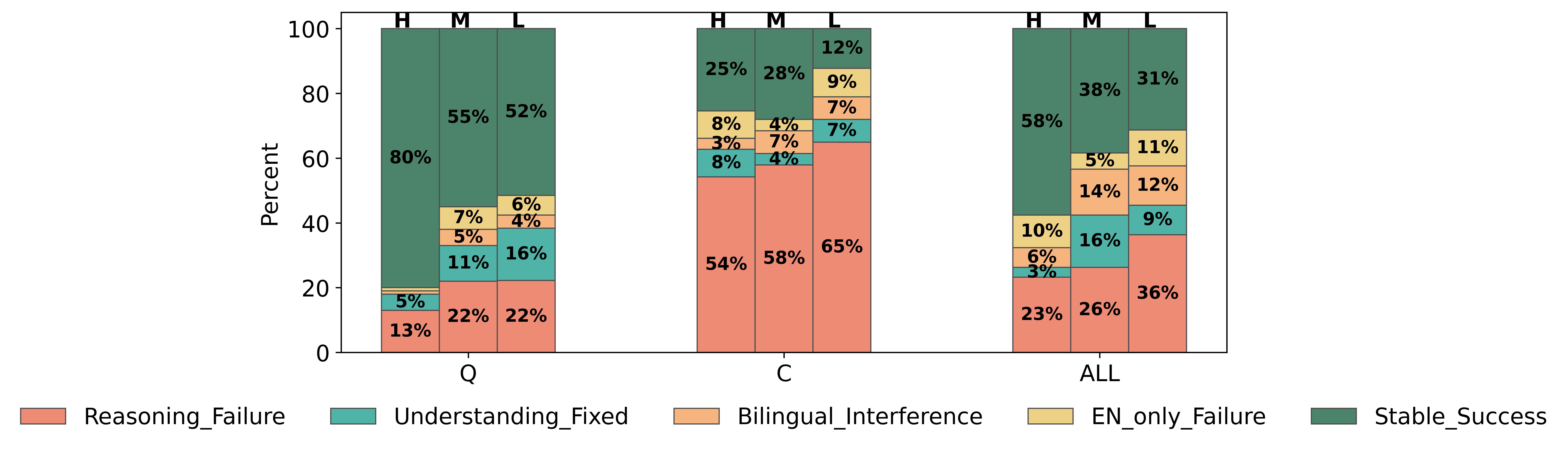}
    \caption{Root-cause composition by resource tier (H/M/L) across text scopes for MMLU-ProX-Lite, LLaMA. Bars show percent of outcomes per tier with labels; colors denote failure categories and stable success.}
    \label{fig:QLRC}
    \end{figure*}
    \begin{figure*}[!t]
    \centering
    \includegraphics[width=\linewidth]{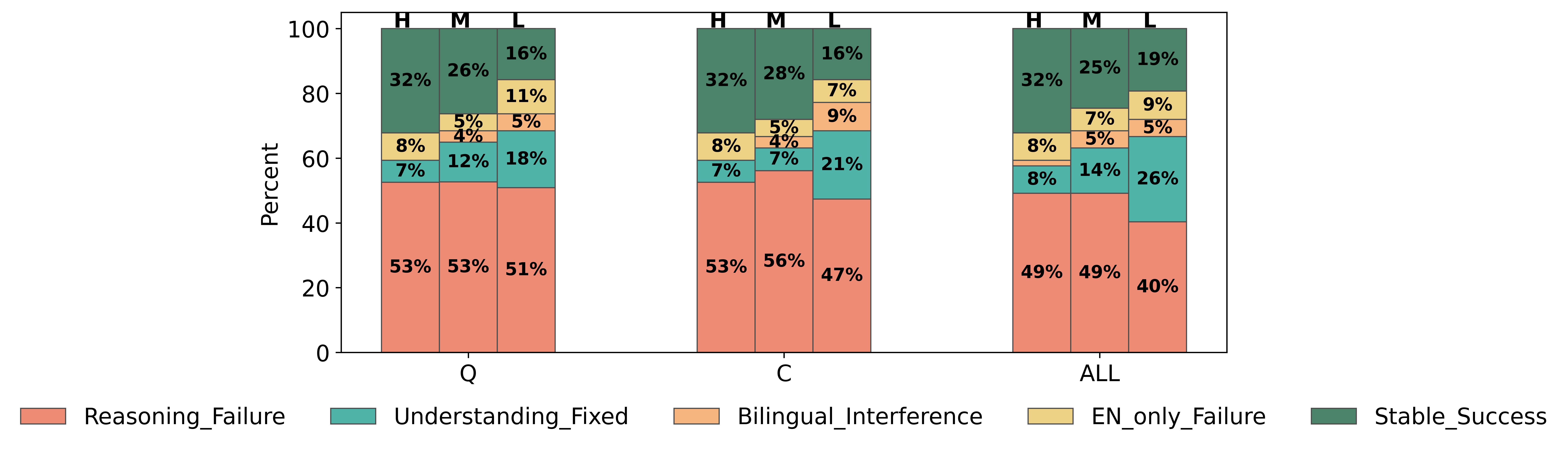}
    \caption{Root-cause composition by resource tier (H/M/L) across text scopes for MMLU-ProX-Lite, Qwen. Bars show percent of outcomes per tier with labels; colors denote failure categories and stable success.}
    \label{fig:QMRC}

\end{figure*}
    \begin{figure*}
    \centering
    \includegraphics[width=\linewidth]{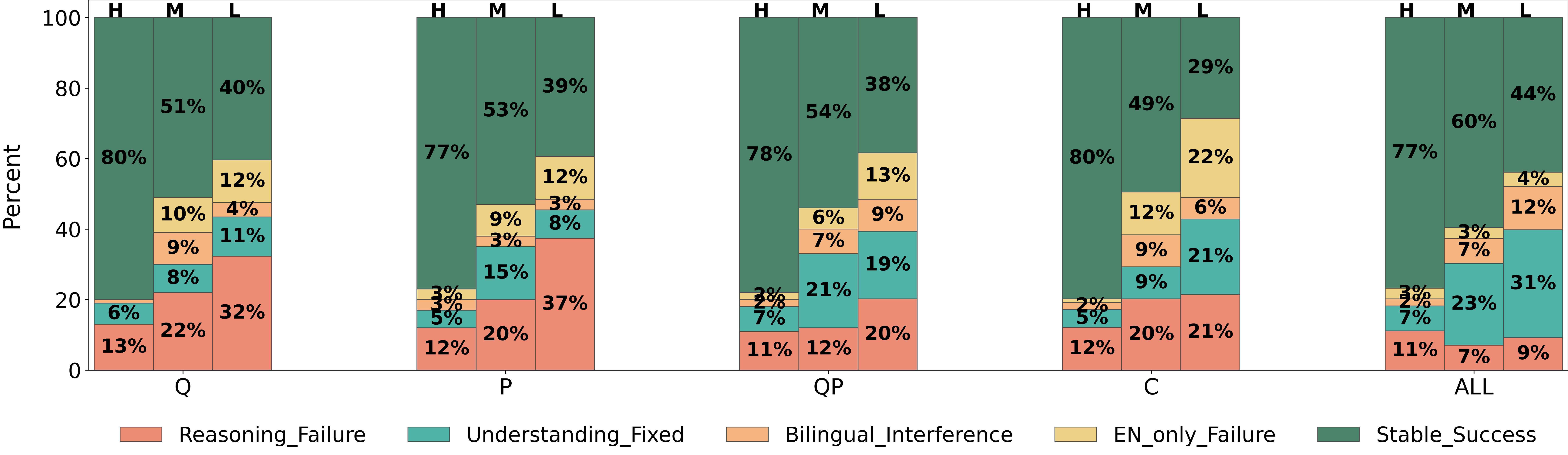}
    \caption{Root-cause composition by resource tier (H/M/L) across text scopes for Belebele, Qwen. Bars show percent of outcomes per tier with labels; colors denote failure categories and stable success.}
  \label{fig:QBRC}
\end{figure*}

\section{Robustness, Sensitivity, and Cost}
\label{app:robustness}

\subsection{Statistical Stability}
\label{app:stability}
For method $m$ and the target-only baseline T, let $\hat p_m$ and $\hat p_T$ denote pooled example-level accuracies in a resource tier containing $N$ examples. We report
\begin{align}
\widehat{\Delta}
&=100(\hat p_m-\hat p_T),\\
\widehat{\mathrm{SE}}
&=100\sqrt{
\frac{\hat p_m(1-\hat p_m)}{N}
+\frac{\hat p_T(1-\hat p_T)}{N}},\\
\mathrm{CI}_{95\%}
&=\widehat{\Delta}\pm1.96\,\widehat{\mathrm{SE}}.
\end{align}
Table~\ref{tab:stability} reports the resulting intervals. The Low tier contains Urdu and Yoruba; its intervals therefore characterize stability over the evaluated examples in these two languages.

\begin{table*}[!t]
\centering
\small
\setlength{\tabcolsep}{3.2pt}
\resizebox{\textwidth}{!}{%
\begin{tabular}{lllr r ll}
\toprule
Model & Dataset & Tier & \#Langs & $N$ & RouteGate $\Delta$Acc vs. T [95\% CI] & SeqGate $\Delta$Acc vs. T [95\% CI]\\
\midrule
LLaMA & Belebele & High & 4 & 317 & +2.32 [-3.60, +8.24] & +1.32 [-4.67, +7.31]\\
LLaMA & Belebele & Mid & 3 & 239 & +4.31 [-3.81, +12.43] & -1.20 [-9.54, +7.14]\\
LLaMA & Belebele & Low & 2 & 158 & +10.60 [-0.36, +21.56] & \textbf{+25.11 [+14.55, +35.67]}\\
LLaMA & MMLU-ProX-Lite & High & 4 & 189 & +1.27 [-8.36, +10.90] & -0.89 [-10.45, +8.67]\\
LLaMA & MMLU-ProX-Lite & Mid & 3 & 146 & +2.31 [-7.85, +12.47] & +8.09 [-2.35, +18.53]\\
LLaMA & MMLU-ProX-Lite & Low & 2 & 92 & +5.15 [-7.51, +17.81] & +10.65 [-2.34, +23.64]\\
Qwen & Belebele & High & 4 & 317 & +3.49 [-2.48, +9.46] & \textbf{+8.26 [+2.65, +13.87]}\\
Qwen & Belebele & Mid & 3 & 239 & +3.05 [-5.32, +11.42] & -3.32 [-11.89, +5.25]\\
Qwen & Belebele & Low & 2 & 158 & \textbf{+25.19 [+14.69, +35.69]} & \textbf{+37.69 [+28.01, +47.37]}\\
Qwen & MMLU-ProX-Lite & High & 4 & 189 & +4.26 [-5.60, +14.12] & +9.35 [-0.57, +19.27]\\
Qwen & MMLU-ProX-Lite & Mid & 3 & 146 & +9.11 [-1.93, +20.15] & \textbf{+18.97 [+7.85, +30.09]}\\
Qwen & MMLU-ProX-Lite & Low & 2 & 92 & +9.47 [-4.08, +23.02] & \textbf{+22.79 [+9.02, +36.56]}\\
\bottomrule
\end{tabular}
}%
\caption{Accuracy differences relative to T with 95\% normal-approximation confidence intervals. Bold intervals exclude zero.}
\label{tab:stability}
\end{table*}

\subsection{Risk Index Ablation and Routing-Budget Sensitivity}
\label{app:ri-robustness}

\paragraph{Factor ablation.}
Table~\ref{tab:ri-ablation} evaluates the components of the Risk Index within high-, mid-, and low-resource language--scope groups. Each variant is assessed using Spearman correlation with target-only error and AUROC for identifying groups above the median target-only error. We write
\begin{equation}
\mathrm{RI}=\mathrm{RF}\times\text{ECE factor}\times\text{Entropy factor},
\label{eq:risk-index}
\end{equation}
where the entropy factor is the normalized low-entropy certainty term, not raw entropy. No single variant dominates every metric; the full product provides a conservative conjunction of failure prevalence, miscalibration, and confident prediction.

\begin{table*}[!t]
\centering
\small
\resizebox{\textwidth}{!}{%
\begin{tabular}{lccccc}
\toprule
Variant & \makecell{High-Res.\\err/AUROC} & \makecell{Mid-Res.\\err/AUROC} & \makecell{Low-Res.\\err/AUROC} & \makecell{err\\average} & \makecell{AUROC\\average}\\
\midrule
RF only & 0.43 / 0.94 & 0.57 / 0.86 & $-0.15$ / 0.32 & 0.28 & 0.71\\
ECE only & 0.48 / 0.91 & 0.27 / 0.68 & \textbf{0.80 / 1.00} & 0.51 & \textbf{0.86}\\
Entropy only & \textbf{0.62} / 0.91 & 0.45 / 0.79 & 0.58 / 0.76 & \textbf{0.55} & 0.82\\
RF $\times$ ECE & 0.49 / 0.96 & 0.40 / 0.79 & 0.42 / 0.68 & 0.44 & 0.81\\
RF $\times$ Entropy & \textbf{0.62 / 1.00} & \textbf{0.64 / 0.89} & $-0.07$ / 0.36 & 0.40 & 0.75\\
ECE $\times$ Entropy & 0.56 / 0.91 & 0.31 / 0.71 & 0.56 / 0.84 & 0.48 & 0.82\\
\textbf{RI = RF $\times$ ECE $\times$ Entropy (Ours)} & 0.61 / 0.98 & 0.37 / 0.77 & 0.51 / 0.72 & 0.50 & 0.82\\
\bottomrule
\end{tabular}
}%
\caption{Ablation study of the Risk Index for LLaMA on Belebele. Each resource-tier entry reports Spearman correlation with target-only error / AUROC for detecting groups with above-median target-only error.}
\label{tab:ri-ablation}
\end{table*}

\begin{table*}[!t]
\centering
\small
\resizebox{\textwidth}{!}{%
\begin{tabular}{lcccc}
\toprule
Model / Dataset & Top 25\% & Top 50\% & Top 75\% & Top 100\%\\
\midrule
LLaMA / Belebele & +3.27 / $-0.07$ & +6.54 / $-0.34$ & +10.73 / $-1.00$ & +10.93 / $-0.71$\\
LLaMA / MMLU-ProX-Lite & +1.27 / +0.63 & +1.45 / +0.11 & +4.45 / +0.54 & +4.41 / +0.27\\
Qwen / Belebele & +2.12 / +1.22 & +3.62 / +0.86 & +5.44 / +0.52 & +5.43 / +0.87\\
Qwen / MMLU-ProX-Lite & +0.80 / +1.99 & +0.92 / +2.27 & +1.43 / +1.94 & +10.54 / +1.49\\
\bottomrule
\end{tabular}
}%
\caption{SeqGate Risk Index threshold sensitivity. Each cell reports $\Delta$Acc / $\Delta$ECE versus T.}
\label{tab:ri-threshold}
\end{table*}

\begin{table*}[!t]
\centering
\scriptsize
\setlength{\tabcolsep}{2pt}
\resizebox{\textwidth}{!}{%
\begin{tabular}{lllrrrrr}
\toprule
Model & Dataset & Tier & RI threshold & RouteGate routed & SeqGate routed & RouteGate cost ($\times$T) & SeqGate cost ($\times$T)\\
\midrule
LLaMA & Belebele & High & 0.500 & 19.8\% & 35.1\% & 1.35 & 1.75\\
LLaMA & Belebele & Mid & 0.315 & 37.8\% & 55.0\% & 1.62 & 2.00\\
LLaMA & Belebele & Low & 0.115 & 83.7\% & 90.3\% & 2.43 & 2.00\\
LLaMA & MMLU-ProX-Lite & High & 0.500 & 3.3\% & 40.0\% & 1.03 & 2.00\\
LLaMA & MMLU-ProX-Lite & Mid & 0.315 & 15.5\% & 60.0\% & 1.04 & 2.00\\
LLaMA & MMLU-ProX-Lite & Low & 0.115 & 35.8\% & 89.0\% & 1.06 & 2.00\\
Qwen & Belebele & High & 0.500 & 13.0\% & 30.1\% & 1.20 & 1.30\\
Qwen & Belebele & Mid & 0.315 & 30.4\% & 55.0\% & 1.46 & 2.00\\
Qwen & Belebele & Low & 0.115 & 69.2\% & 80.0\% & 2.14 & 2.00\\
Qwen & MMLU-ProX-Lite & High & 0.500 & 11.9\% & 41.2\% & 1.02 & 2.00\\
Qwen & MMLU-ProX-Lite & Mid & 0.315 & 26.6\% & 70.0\% & 1.07 & 2.00\\
Qwen & MMLU-ProX-Lite & Low & 0.115 & 45.1\% & 80.1\% & 1.05 & 2.00\\
\bottomrule
\end{tabular}
}%
\caption{Tier-level routing rates and normalized LLM inference cost relative to T. Translation-system cost and latency are not included.}
\label{tab:cost}
\end{table*}

\paragraph{Routing-budget sensitivity.}
Table~\ref{tab:ri-threshold} varies the fraction of highest-RI language--scope groups routed to \textsc{SeqGate}. Entries are changes in accuracy and ECE relative to T, in percentage points. Broader routing often improves accuracy, although the relationship is dataset-dependent and not monotonic in every setting.

\paragraph{\textsc{RouteGate} threshold protocol.}
The low-ECE layers in Equation~\ref{eq:ece-layer} and the trigger thresholds in Equation~\ref{eq:route-trigger} are selected on development data and frozen before test evaluation. ECE is used only for offline layer selection; each online decision uses per-example confidence and final-versus-ensemble disagreement. The answer-scoring rule is unchanged.

\subsection{RAAI Efficiency and Inference Cost}
\label{app:cost}
Table~\ref{tab:cost} reports routing rates and normalized LLM inference cost relative to T. The reported cost reflects the implementation's processing cost for longer bilingual or sequential prompts and excludes external machine-translation cost. Deployment with EN, TEN, or SEQ would require cached English fields or a runtime translation system.

\section{Disclosure of AI Tools}
\label{app:reproducibility}
Experiments were run with fixed model, prompt, calibration, and routing configurations documented in the paper and accompanying code. AI tools were used for coding assistance (OpenAI Codex), language editing, and related-work search (OpenAI ChatGPT and Perplexity). The authors verified the resulting code, citations, analyses, and conclusions.

\end{document}